\documentclass{article} 
\usepackage{nips13submit_e,times}
\usepackage{hyperref}
\usepackage{url}
\usepackage{epsf}
\usepackage{epsfig}

\usepackage[numbers]{natbib}
\usepackage{setspace}

\usepackage{bm}
\usepackage{amssymb}
\usepackage{amsmath}
\usepackage{algorithm}
\usepackage{algorithmic}
\usepackage{graphicx}

\def\bmu{{\bm \mu}}

\def\bSigma{{\bm \Sigma}}

\def\bxi{{\bm \xi}}

\def\bdelta {{\bm \delta}}
\def\bzero {{\bm 0}}

\def \btheta {\bm \theta}

\def \bphi {\bm \phi} 
\def \bxi {\bm \xi} 

\newtheorem{theorem}{Theorem}

\long\def\ignore#1{}

\title{Stochastic Bound Majorization}

\author{
Anna Choromanska \\
Department of Electrical Engineering\\
Columbia University\\
New York\\
\texttt{aec2163@columbia.edu} \\
\And
Tony Jebara \\
Department of Computer Science\\
Columbia University\\
New York\\
\texttt{jebara@cs.columbia.edu} \\
}
\nipsfinalcopy

\begin{document}

\maketitle

\begin{abstract}
  Recently a majorization method for optimizing partition functions of
  log-linear models was proposed alongside a novel quadratic
  variational upper-bound. In the batch setting, it outperformed
  state-of-the-art first- and second-order optimization methods on
  various learning tasks. We propose a stochastic version of this
  bound majorization method as well as a low-rank modification for
  high-dimensional data-sets. The resulting stochastic second-order
  method outperforms stochastic gradient descent (across variations
  and various tunings) both in terms of the number of iterations and
  computation time till convergence while finding a better quality
  parameter setting. The proposed method bridges first- and
  second-order stochastic optimization methods by maintaining a
  computational complexity that is linear in the data dimension and
  while exploiting second order information about the pseudo-global
  curvature of the objective function (as opposed to the local
  curvature in the Hessian).
\end{abstract}

\section{Introduction}
Stochastic learning algorithms are of central interest in machine
learning due to their simplicity and, as opposed to batch methods,
their low memory and computational complexity requirements
\cite{bottou-98x, Littlestone:1988:LQI:639961.639994,
  rosenblatt58a}. For instance, stochastic learning algorithms are
commonly used to train deep belief networks
(DBNs)~\cite{Bottou_stochasticgradient, lecun-98x} which perform
extremely well on tasks involving massive data-sets
\cite{DBLP:conf/nips/KrizhevskySH12,
  DBLP:conf/interspeech/KingsburySS12}. Stochastic algorithms are
also broadly used to train Conditional Random Fields (CRFs)
\cite{Vishwanathan:2006:ATC:1143844.1143966}, solve maximum
likelihood problems \cite{RouxF10} and perform variational inference
\cite{DBLP:conf/nips/WangB12}.

Most stochastic optimization approaches fall into two groups:
first-order methods and second-order methods. Popular first-order
methods include stochastic gradient descent (SGD)
\cite{ROBINS-MONRO51} and its many extensions
\cite{DBLP:conf/nips/RouxSB12, Polyak:1992:ASA:131092.131098,
  Tseng:1998:IGM:588881.588930, DBLP:journals/mp/Nesterov09, Kesten,
  DBLP:journals/siamjo/BlattHG07}. These methods typically have low
computational cost per iteration (such as $\mathcal{O}(d)$ where $d$
is the data dimensionality) and either sub-linear (most stochastic
gradient methods) or linear (as shown recently
in~\cite{DBLP:conf/nips/RouxSB12}) convergence rate which makes them
particularly relevant for large-scale learning. Despite its
simplicity, SGD has many drawbacks: it has slow asymptotic convergence
to the optimum \cite{Vishwanathan:2006:ATC:1143844.1143966}, it has
limited ability to handle certain regularized learning problems such
as $l_1$-regularization \cite{Xiao09dualaveraging}, it requires step-size
tuning and it is difficult to parallelize
\cite{DBLP:conf/icml/LeNCLPN11}. Many works have tried to incorporate
second-order information (i.e. Hessian) into the optimization problem
to improve the performance of traditional SGD methods. A
straightforward way of doing so is to simply replace the gain in SGD
with the inverse of the Hessian matrix which, when naively
implemented, induces a computational complexity of
$\mathcal{O}(d^3)$. This makes the approach impractical for large
problems. The trade-offs in large-scale learning for various
prototypical batch and stochastic learning algorithms are conveniently
summarized in~\cite{DBLP:conf/nips/BottouB07}). Therein, several new
methods are developed, including variants of Newton's method which use both gradient and Hessian information to compute the descent
direction. By carefully exploring different first- and second-order
techniques, the overall computational complexity of optimization can
be reduced as in the Stochastic Meta-Descent (SMD) algorithm
\cite{Schraudolph99localgain}. Although it still uses the gradient
direction to converge, SMD also efficiently exploits certain
Hessian-vector products to adapt the gradient step-size. The algorithm
is shown to converge to the same quality solution as limited-memory
BFGS (LBFGS) an order of magnitude faster for CRF training
\cite{Vishwanathan:2006:ATC:1143844.1143966}. There also exists
stochastic versions of quasi-Newton methods like online BFGS and
online LBFGS \cite{Schraudolph07astochastic}, the latter applicable to
large-scale problems, which while using convenient size mini-batches
performs comparably to a well-tuned natural gradient descent
\cite{Amari:2000:AMR:1121517.1121530} on the task of training CRFs,
but at the same time is more scalable. In each iteration the inverse
of the Hessian, that is assumed to have no negative eigenvalues, is
estimated. Computational complexity of this online LBFGS method is
$\mathcal{O}(md)$ per iteration, where $m$ is the size of the buffer
used to estimate the inverse of the curvature. The method degrades
(large $m$) for sparse data-sets. Another second-order stochastic
optimization approach proposed in the literature explores diagonal
approximations of the Hessian matrix or Gauss-Newton
matrix\cite{lecun-98x, Bordes:2009:SCQ:1577069.1755842}. In some cases
this approach appears to be overly simplistic
\cite{DBLP:conf/icml/Martens10}, but turned out successful in very
particular applications, i.e. for learning with linear Support Vector
Machines \cite{Bordes:2009:SCQ:1577069.1755842}. There is also a large
body of work on stochastic second-order methods particularly
successful in training deep belief network like Hessian-free
optimization \cite{DBLP:conf/icml/Martens10}. Finally, there are also
many hybrid methods using existing stochastic optimization tools as
building blocks and merging them to obtain faster and more robust
learning algorithms \cite{RouxF10, journals/siamsc/FriedlanderS12}.
 
This paper contributes to the family of existing second-order
stochastic optimization methods with a new algorithm that is using a globally
guaranteed quadratic bound with a curvature different than the
Hessian. Therefore our approach is not merely a variant of Newton's
method. This is a stochastic version of a recently proposed
majorization method \cite{JebCho12} which performed maximum (latent)
conditional likelihood problems more efficiently than other
state-of-the-art first- and second- order batch optimization methods
like BFGS, LBFGS, steepest descent (SD), conjugate gradient (CG) and
Newton. The corresponding stochastic bound majorization method is
compared with a well-tuned SGD with either constant or adaptive gain
in $l_2$-regularized logistic regression and turns out to outperform
competitor methods in terms of the number of iterations, the
convergence time and even the quality of the obtained solution
measured by the test error and test likelihood.

\section{Preliminaries}
A staggering number of machine learning and statistics frameworks
involve linear combinations or cascaded linear combinations of
soft-maximum functions:
\[s(\btheta) = \sum_{j=1}^{t}\gamma_i\log\sum_{y}h(y)\exp(\btheta^{\top}{\bf f}(y)),
\]
where $\btheta \in \mathbb{R}^{d}$ is a parameter vector, ${\bf f}:
\Omega \rightarrow \mathbb{R}^d$ is any vector-valued function mapping
an input $y$ to some arbitrary vector (we assume $\Omega$ is finite
and $|\Omega| = n$ is enumerable), $t$ is the size of the data-set and
$\gamma_i$ is some non-negative weight. These functions emerge in
multi-class logistic regression, CRFs
\cite{DBLP:conf/icml/LaffertyMP01}, hidden variable problems, DBNs
\cite{DBLP:journals/neco/HintonOT06}, discriminatively trained speech
recognizers \cite{Bahl96discriminativetraining} and maximum entropy
problems \cite{berger97improved}. For simplicity, we focus herein on 
the CRF training problem in particular. CRFs use the density model:
\[p(y|x_j,\btheta) = \frac{1}{Z_{x_j}(\btheta)}h_{x_j}(y)\exp(\btheta^{\top}{\bf f}_{x_j}(y)),
\]
where $\{(x_1,y_1), \dots, (x_t,y_t)\}$ are {\em iid} input-output
pairs and $Z_{x_j}(\btheta)$ is a partition function:
$Z_{x_j}(\btheta) = \sum_{y \in
  \Omega_j}h_{x_j}(y)\exp(\btheta^{\top}{\bf f}_{x_j}(y))$. Following
the maximum likelihood approach, the objective function to maximize in
this setting is:
\begin{equation}
J(\btheta) = \sum_{j = 1}^{t}\left[\log\frac{h_{x_j}(y_j)}{Z_{x_j}(\btheta)} + \btheta^{\top}{\bf f}_{x_j}(y_j)\right] - \frac{\lambda}{2}\|\btheta\|^2,
\label{eq:objective}
\end{equation}
where $\lambda$ is a regularization hyper-parameter. Let $J(\btheta) =
\sum_{j=1}^{t}J_j(\btheta)$, where $J_j(\btheta) =
\log\frac{h_{x_j}(y_j)}{Z_{x_j}(\btheta)} + \btheta^{\top}{\bf
  f}_{x_j}(y_j) - \frac{\lambda}{2t}\|\btheta\|^2$. For large numbers
of data points $t$, and potentially large dimensionality $d$,
summations in Equation~\ref{eq:objective} need not be handled in a
batch form, but rather, can be processed stochastically or
semi-stochastically. We next review the most commonly used
stochastic algorithm, SGD, which will be a key comparator
for our stochastic bound majorization algorithm.

\subsection{Stochastic gradient descent methods}
Batch gradient descent updates the parameter vector $\btheta$ after seeing the entire training data-set using the following formula:
\[\btheta = \btheta - \eta\bmu(\btheta) = \btheta - \eta\sum_{j=1}^{t}\bmu_j(\btheta),
\] 
where $\bmu(\btheta) = \bigtriangledown_{\btheta} J(\btheta)$,
$\bmu_j(\btheta) = \bigtriangledown J_j(\btheta)$ and $\eta$ is
typically chosen via line search. In contrast, stochastic gradient
descent updates the parameter vector $\btheta$ after seeing each
training data point (resp. each mini-batch of data points) as follows:
\[\btheta_{i+1} = \btheta_i -
\eta_i\bmu_j(\btheta_i),\:\:\:\:\:\:\:\:\text{resp.}\:\:\btheta_{i+1}
= \btheta_i - \eta_i\sum_{j=1}^{m}\bmu_j(\btheta_i),
\] 
where $\btheta_i$ is the current parameter vector, $\eta_i$ is the
current gain or step-size, and $m<<t$ is the size of the mini-batch.
We assume a data point is randomly selected and take its index to be
$j$. We intentionally index $\btheta$ with $i$ and $i+1$ to emphasize
that the update is done after seeing single example (resp. mini-batch
of examples). For the batch method, we do not index $\btheta$ since
the update is done after passing through the entire data-set
(epoch). In the experimental section we explore two existing variants
of stochastic gradient descent: one with constant gain (SGD) and one
with adaptive gain (ASGD). For the latter, we consider two strategies
for modifying the gain and present the results for the better one (in Section~\ref{sec:Experiments} the
absence of a $\tau$ value indicates that the second strategy is better since it
does not require a $\tau$ parameter):
\begin{itemize}
\item{$\eta_i = \frac{\tau}{\tau + i}\eta_0$}
\item{$\eta_i = \frac{\eta_0}{i}$, \:\:\:\:\:\:\:\:\:\:\:\:\:\:\:\:\:\:\:\:\: where $\eta_0, \tau > 0$ \: are tuning parameters.}
\end{itemize}

\subsection{Batch bound majorization algorithm}
\begin{algorithm}
\begin{tabular}{ll}
\setlength\tabcolsep{0pt}
\hspace{-0.18in}
\mbox{
\begin{tabular}{l}
\vspace{-1.03in} \\
\hline
Input Parameters $\tilde{\btheta}, {\bf f}(y), h(y)  \: \forall y \in \Omega $ \\
\hline
Init $z \rightarrow 0^+, {\bf g} = \bzero, \bSigma=z {\bf I} \:\: $\\
For each $y \in \Omega$  \{ \\
$\:\:\:\:\:\:\:\:\:\:\:\: \alpha = h(y) \exp(\tilde{\btheta}^\top {\bf f}(y)); \:\:{\bf l} = {\bf f}(y)-{\bf g} $ \\
$\:\:\:\:\:\:\:\:\:\:\:\: \beta = \frac{\tanh(\frac{1}{2} \log (\alpha/z))}{2 \log ( \alpha/z)}; \:\:\kappa = \frac{\alpha}{z+\alpha} $ \\
$\:\:\:\:\:\:\:\:\:\:\:\: \bSigma\:\:\:\!\!\!\;  \!\;+\!= \beta {\bf l}{\bf l}^\top$\\
$\:\:\:\:\:\:\:\:\:\:\:\: {\bf g}\:\:\!\!\:\; \: +\!= \kappa {\bf l} $ \\
$\:\:\:\:\:\:\:\:\:\:\:\: z\:\;  \:\: +\!= \alpha \:\:\:\:\:\:\: \} $  \\
\hline
Output $z,{\bf g},\bSigma$ \\
\hline
\vspace{0.438in} 
\end{tabular}
}
& \hspace{0.9in}  \epsfysize=1.5in \epsfbox{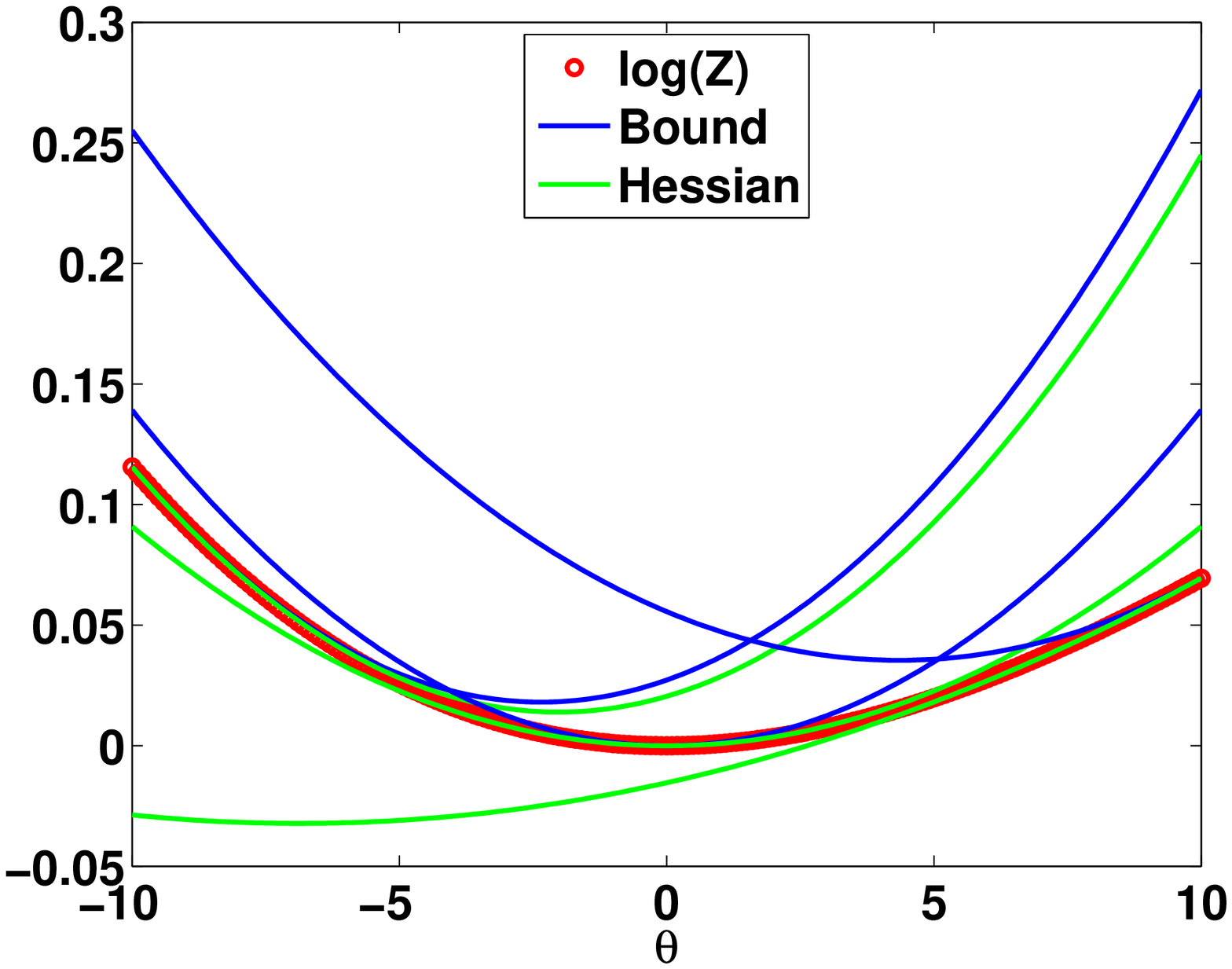} 
\end{tabular}
\vspace{-0.64in}
\caption{\textsf{Batch Bound}}
\label{alg:batchbound}
\end{algorithm}

\begin{theorem}
Algorithm~\ref{alg:batchbound} finds $z, {\bf g}, \bSigma$ such that
$z \exp  ( \frac{1}{2}
  (\btheta-\tilde{\btheta})^\top \bSigma (\btheta-\tilde{\btheta}) +
  (\btheta-\tilde{\btheta})^\top\! {\bf g}  )$ upper-bounds
$Z(\btheta)=\sum_y h(y) \exp(\btheta^\top {\bf f}(y))$
for any
$\btheta,\tilde{\btheta}, {\bf f}(y) \in \mathbb{R}^d$ and $h(y) \in
\mathbb{R}^+$ for all $y \in \Omega$.
\label{thm:main}
\end{theorem}
The stochastic bound majorization algorithm proposed in this paper is
a stochastic variant of the batch method described in
\cite{JebCho12}. The bound is depicted in Theorem~\ref{thm:main}. The
figure near Algorithm~\ref{alg:batchbound} shows typical bounds the
algorithm recovers (in blue) and also shows (in green) examples what
happens when the bound's $\bSigma$ matrix is replaced by the Hessian
matrix which yields a second-order approximation to the
function. These tight quadratic bounds facilitate majorization:
solving an optimization problem by iteratively finding the optima of
simple bounds on it \cite{LeeuwHeiser2006} as popularized by the
Expectation-Maximization (EM) algorithm
\cite{Dempster77maximumlikelihood}. Majorization was shown to achieve
faster and monotonically convergent performance in CRF learning and
maximum latent conditional likelihood problems with a clear advantage
over state-of-the-art first- and second-order batch methods
\cite{JebCho12}. The batch bound majorization algorithm applied
to Equation~\ref{eq:objective} updates the parameter
vector $\btheta$ after seeing the entire training data-set using the
following formula:
\begin{equation}
\btheta = \btheta - \eta\bSigma^{-1}(\btheta)\mu(\btheta),
\label{eq:batchupdate}
\end{equation} 
where $\bSigma(\btheta) = \sum_{j=1}^t\bSigma_j(\btheta) + \lambda{\bf
  I}$, $\bmu(\btheta) = \sum_{j=1}^t\bmu_j(\btheta) =
\sum_{j=1}^t({\bf g}_j(\btheta) - {\bf f}_j(y_j)) +
\lambda\btheta$. Here, each $\bSigma_j$ and ${\bf g}_j$ is computed
using Algorithm~\ref{alg:batchbound} (for details see~\cite{JebCho12})
and $\eta \in (0,2)$ guarantees monotonic
improvement~\cite{DBLP:conf/icml/SalakhutdinovR03} (typically we
set $\eta = 1$).

\section{Stochastic bound majorization algorithm}
The intuition behind stochastic gradient descent is to compute the
gradient over a single representative data-point rather than an
iterative computation (namely a summation) over all data-points. This
allows us to interleave parameter updates into the gradient
computation rather than wait for it to terminate after a full
epoch. We herewith extend this intuition into a bound majorization
setting. Unfortunately, the update rule for majorization involves a
matrix inverse rather than a simple sum across the data. We will
re-cast this update rule as an iterative computation over the data-set
which will then admit a stochastic incarnation.

\subsection{Full-rank version}
Notice that the batch update in Equation~\ref{eq:batchupdate} is the solution of the linear system:
\[\left( \sum_{j=1}^t\bSigma_j + \lambda{\bf I} \right)\bdelta = \sum_{j=1}^t\bmu_j
\]
For the ease of notation we denote $\bSigma_j(\btheta)$ as $\bSigma_j$
and $\bmu_j(\btheta)$ as $\bmu_j$. Rewrite the linear system as
$\bSigma\bdelta = {\bf u}$. We then define the inverse matrix ${\bf M}
= \bSigma^{-1}$ and write the solution as $\bdelta = {\bf M}{\bf
  u}$. Clearly, the matrix inversion cannot be performed each time we
process a single data-point in a stochastic setting since a
computational complexity of $\mathcal{O}(d^3)$ is prohibitive in
a stochastic setting.  Therefore, consider an online version of
Algorithm~\ref{alg:batchbound} which computes the matrix inversion of
$\bSigma$ incrementally using the Sherman-Morrison formula $(\bSigma +
{\bf q}_i{\bf q}_i^{\top})^{-1} = \bSigma^{-1} - (\bSigma^{-1}{\bf
  q}_i{\bf q}_i^{\top}\bSigma^{-1})/(1 + {\bf
  q}_i^{\top}\bSigma^{-1}{\bf q}_i)$. Sherman-Morrison works with the
inverse matrix ${\bf M} = \bSigma^{-1}$ instead. Initialize ${\bf M}_0
= \frac{1}{\lambda}{\bf I}$ and incrementally increase ${\bf M}$ using
the update rule:
\[{\bf M}_{i+1} = {\bf M}_i - \frac{{\bf M}_i{\bf q}_i{\bf q}_i^{\top}{\bf M}_i}{1 + {\bf q}_i^{\top}{\bf M}_i{\bf q}_i}, \:\:\:\:\:\text{where}\:\:\:{\bf q}_i = \sqrt{\beta_i}({\bf f}_i - {\bf g}_i) = \sqrt{\beta_i}{\bf l}_i,
\]
where the index $i$ ranges over all rank $1$ updates to the matrix
$\bSigma$ for all elements of $\Omega$ as well as $j = 1, \dots,
t$. Finally, the solution is obtained by multiplying ${\bf M}$ with
${\bf u}$. This avoids $\mathcal{O}(d^3)$ inversion and solves the
linear system in $\mathcal{O}(tnd^2)$. Let $T = tn$. We will now
reformulate the batch update from Equation~\ref{eq:batchupdate}
by using the Sherman-Morrison technique. For the ease of notation,
introduce $\bxi$'s such that $\forall_{i = 1, 2, \dots, T}\bmu_i =
\bmu_{i-1} + \bxi_{i-1}$ (thus $\bxi_{i-1} = \kappa_{i-1}{\bm l}_{i-1}
- {\bm f}_{i-1} + \lambda\btheta/t$).
\[\btheta = \btheta - \eta{\bm M_T}\bmu_T = \btheta - \eta\left[{\bm M}_{T-1} - \frac{{\bm M}_{T-1}{\bf q}_{T-1}{\bf q}_{T-1}^{\top}{\bm M}_{T-1}}{1 + {\bf q}_{T-1}^{\top}{\bf M}_{T-1}{\bf q}_{T-1}}\right](\bmu_{T-1} + \bxi_{T-1}) 
\]
\[= \btheta - \eta{\bm M}_{T-1}\bmu_{T-1} - \eta{\bm M}_{T-1}\bxi_{T-1}+  \frac{{\bm M}_{T-1}{\bf q}_{T-1}{\bf q}_{T-1}^{\top}{\bm M}_{T-1}}{1 + {\bf q}_{T-1}^{\top}{\bf M}_{T-1}{\bf q}_{T-1}} \bmu_{T-1}
\]
\[ + \frac{{\bm M}_{T-1}{\bf q}_{T-1}{\bf q}_{T-1}^{\top}{\bm M}_{T-1}}{1 + {\bf q}_{T-1}^{\top}{\bf M}_{T-1}{\bf q}_{T-1}}\bxi_{T-1}.
\]
We can further expand ${\bm M}_{T-1}\bmu_{T-1}$ as:
\[{\bm M}_{T-1}\bmu_{T-1} = \left[{\bm M}_{T-2} -\frac{{\bm M}_{T-2}{\bf q}_{T-2}{\bf q}_{T-2}^{\top}{\bm M}_{T-2}}{1 + {\bf q}_{T-2}^{\top}{\bf M}_{T-2}{\bf q}_{T-2}}\right](\bmu_{T-2} + \bxi_{T-2})
\]
\[= {\bm M}_{T-2}\bmu_{T-2} + {\bm M}_{T-2}\bxi_{T-2} - \frac{{\bm M}_{T-2}{\bf q}_{T-2}{\bf q}_{T-2}^{\top}{\bm M}_{T-2}}{1 + {\bf q}_{T-2}^{\top}{\bf M}_{T-2}{\bf q}_{T-2}} \bmu_{T-2} -\frac{{\bm M}_{T-2}{\bf q}_{T-2}{\bf q}_{T-2}^{\top}{\bm M}_{T-2}}{1 + {\bf q}_{T-2}^{\top}{\bf M}_{T-2}{\bf q}_{T-2}}\bxi_{T-2}
\]
and again we can further expand ${\bm M}_{T-2}\bmu_{T-2}$ as: 
\[{\bm M}_{T-2}\bmu_{T-2} = \left[{\bm M}_{T-3} - \frac{{\bm M}_{T-3}{\bf q}_{T-3}{\bf q}_{T-3}^{\top}{\bm M}_{T-3}}{1 + {\bf q}_{T-3}^{\top}{\bf M}_{T-3}{\bf q}_{T-3}}\right](\bmu_{T-3} + \bxi_{T-3})
\]
\[= {\bm M}_{T-3}\bmu_{T-3} + {\bm M}_{T-3}\bxi_{T-3} - \frac{{\bm M}_{T-3}{\bf q}_{T-3}{\bf q}_{T-3}^{\top}{\bm M}_{T-3}}{1 + {\bf q}_{T-3}^{\top}{\bf M}_{T-3}{\bf q}_{T-3}} \bmu_{T-3} - \frac{{\bm M}_{T-3}{\bf q}_{T-3}{\bf q}_{T-3}^{\top}{\bm M}_{T-3}}{1 + {\bf q}_{T-3}^{\top}{\bf M}_{T-3}{\bf q}_{T-3}}\bxi_{T-3}.
\]
We repeat these steps. In the last step we expand ${\bm M}_{1}\bmu_{1}$. We can combine these results and write the following update rule:
\[\btheta = \btheta - \eta{\bm M}_T\bmu_T = \btheta - \eta{\bm M}_0\bmu_0 - \eta\sum_{c = 0}^{T-1}\left[{\bm M}_{c}\bxi_{c} -\frac{{\bm M}_{c}{\bf q}_{c}{\bf q}_{c}^{\top}{\bm M}_{c}}{1 + {\bf q}_{c}^{\top}{\bf M}_{c}{\bf q}_{c}} \bmu_{c} -\frac{{\bm M}_{c}{\bf q}_{c}{\bf q}_{c}^{\top}{\bm M}_{c}}{1 + {\bf q}_{c}^{\top}{\bf M}_{c}{\bf q}_{c}}\bxi_{c}\right]
\]
\[= \btheta - \eta\sum_{c = 0}^{T-1}\left[\left({\bm M}_{c} - \frac{{\bm M}_{c}{\bf q}_{c}{\bf q}_{c}^{\top}{\bm M}_{c}}{1 + {\bf q}_{c}^{\top}{\bf M}_{c}{\bf q}_{c}}\right)\bxi_{c} - \frac{{\bm M}_{c}{\bf q}_{c}{\bf q}_{c}^{\top}{\bm M}_{c}}{1 + {\bf q}_{c}^{\top}{\bf M}_{c}{\bf q}_{c}} \bmu_{c}\right]
\]
The last inequality comes from the fact that $\bmu_0$ is initialized as $\bmu_0 = \bzero$. We have thus rewritten the batch majorization update rule for the parameters as an iterative summation over the data. Analogous to the conversion of a batch gradient descent algorithm into SGD, the most natural way to convert the batch bound majorization algorithm to its stochastic version is to interleave updates of the parameter $\btheta$ rather than waiting for the full summation to terminate (after an epoch) before allowing $\btheta$ to update. This permits us to now write a fully stochastic update rule on the parameter vector $\btheta$ (in original notation):   
\begin{eqnarray*}
\begin{split}
\btheta_{i + 1} = \btheta_i - \eta_i\left[\left({\bm M}_{j}(\btheta_i) -\frac{{\bm M}_{j}(\btheta_i){\bf q}_{j}(\btheta_i){\bf q}_{j}(\btheta_i)^{\top}{\bm M}_{j}(\btheta_i)}{1 + {\bf q}_{j}(\btheta_i)^{\top}{\bf M}_{j}(\btheta_i){\bf q}_{j}(\btheta_i)}\right)\bxi_{j}(\btheta_i)\right.\\
\left. - \frac{{\bm M}_{j}(\btheta_i){\bf q}_{j}(\btheta_i){\bf q}_{j}(\btheta_i)^{\top}{\bm M}_{j}(\btheta_i)}{1 + {\bf q}_{j}(\btheta_i)^{\top}{\bf M}_{j}(\btheta_i){\bf q}_{j}(\btheta_i)} \bmu_{j}(\btheta_i)\right]
\end{split}
\end{eqnarray*}

\begin{algorithm}
\begin{tabular}{ll}
\setlength\tabcolsep{0pt}
\hspace{-0.18in}
\begin{tabular}{l}
Input: $\lambda \in \mathbb{R}^{+}$, $\eta$\\
\hline
Initialize: ${\btheta} \in \mathbb{R}^{d}$, $\bphi = zeros(d)$, ${\bm M} = \frac{1}{\lambda}{\bm I}$, $\bmu = \bzero$\\
\hline
while not converged \{\\
\:\:\:\:\: select data point $j$ \\
\:\:\:\:\: $z \rightarrow 0^+; \:\:{\bm g} = \bzero$\\
\:\:\:\:\:  For each $y \in \Omega$\:\: \{ \\
$\:\:\:\:\:\:\:\:\:\:\:\:\: \alpha = h_j(y)\exp({\btheta}^\top {{\bm f}_j}(y)); \:\:{\bm{l}} = {{\bm f}}_j(y)-{\bm g}; \:\:\beta =\frac{\tanh(\frac{1}{2}\log(\alpha/z)}{2\log(\alpha/z)}; \:\:\kappa = \frac{\alpha}{z + \alpha}\:\:\:\:\:\:\:\:\:\:\:\:\:\:\:\:\:\:\:\:\:\:\:\:\:\:\:\:\:\:$\\
$\:\:\:\:\:\:\:\:\:\:\:\:\: \bxi = \kappa{\bm l} - {\bm f}_j + \lambda\btheta/t$\\
$\:\:\:\:\:\:\:\:\:\:\:\:\: {\bm N}\:\:\:\; = \frac{{\bm M}\beta{\bm l}{\bm l}^{\top}{\bm M}}{1+\beta{\bm l}^{\top}{\bm M}{\bm l}}$\\
$\:\:\:\:\:\:\:\:\:\:\:\:\: {\bm M}\:\!\!\; -\!= {\bm N}$\\
$\:\:\:\:\:\:\:\:\:\:\:\:\: \bphi \:\:\; +\!= {\bm M}\bxi - {\bm N}\bmu$\\
$\:\:\:\:\:\:\:\:\:\:\:\:\: {\bm g} \:\:\:\!\:\!\; +\!= \kappa{\bm l}$\\
$\:\:\:\:\:\:\:\:\:\:\:\:\: \bmu \:\:\; +\!= \bxi$\\
$\:\:\:\:\:\:\:\:\:\:\:\:\:  z\:\;  \:\: +\!= \alpha$ \:\:\} \\
\:\:\:\:\: $\btheta\:\; -\!= \eta\bphi$ \:\:\} \\
\end{tabular}
\end{tabular}
\caption{\textsf{Stochastic Bound}}
\label{alg:stochasticbound}
\end{algorithm}

The stochastic bound majorization algorithm also readily admits
mini-batches. Algorithm~\ref{alg:stochasticbound} captures the
full-rank version of the proposed algorithm (we always use constant
step size $\eta = \frac{1}{t}$).

We have also investigated many other potential variants of
Algorithm~\ref{alg:stochasticbound} including heuristics borrowed from
other stochastic algorithms in the literature
\cite{DBLP:conf/nips/RouxSB12}. Some heuristics involved using memory
to store previous values of updates, gradients and second order matrix
information. Remarkably, all such heuristics and modifications slowed
down the convergence of Algorithm~\ref{alg:stochasticbound}.

On caveat remains. The computational complexity of the proposed
stochastic bound majorization method is $\mathcal{O}(nd^2)$ per
iteration which is less appealing than the $\mathcal{O}(nd)$
complexity of SGD. This shortcoming is resolved in the next
subsection.

\subsection{Low-rank version}
We next provide a low-rank version of
Algorithm~\ref{alg:stochasticbound} to maintain a $\mathcal{O}(nd)$
run-time per stochastic update. Consider the update on $\bphi$ inside
the loop over $y$ in Algorithm~\ref{alg:stochasticbound}. This update
can be rewritten as
\[\bphi +\!= {\bm M}\bxi - {\bm N}\bmu = {\bm M}\bxi - ({\bm M}_{old} - {\bm M}) \bmu = (\bSigma)^{-1}\bxi- ((\bSigma_{old})^{-1} - (\bSigma)^{-1}) \bmu,
\]
where ${\bf M}_{old}$ and $\bSigma_{old}$ are the matrices that, after
being updated, become ${\bf M}$ and $\bSigma$ respectively: ${\bf M} =
{\bf M}_{old} - {\bf N}_{old}$, ${\bf M}_{old} = \bSigma_{old}^{-1}$
and $\bSigma = \bSigma_{old} + {\bf q}_{old}{\bf q}_{old}^{\top} =
\bSigma_{old} + {\beta}_{old}{\bf l}_{old}{\bf l}_{old}^{\top}$ (rank
$1$ update). We can store the matrix $\bSigma$ using a low-rank
representation ${\bf V}^{\top}{\bf S}{\bf V} + {\bf D}$, where $k$ is
a rank ($k << d$), ${\bf V} \in \mathbb{R}^{k \times d}$ is
orthonormal, ${\bf S} \in \mathbb{R}^{k \times k}$ is positive
semi-definite and ${\bf D} \in \mathbb{R}^{d \times d}$ is
non-negative diagonal. We can directly update ${\bf V}$, ${\bf S}$ and
${\bf D}$ online rather than incrementing matrix $\bSigma$ by a rank
$1$ update. In the case of batch bound majorization method this still
guarantees an overall upper bound \cite{JebCho12}. We directly apply
this technique as well in our stochastic setting. Due to space
constraints we will not present this technique (we refer the reader
to~\cite{JebCho12}). Given a low-rank version of the matrix, we use
the Woodbury formula to invert it in each iteration: ${\bf
  \Sigma}^{-1} = {\bf D}^{-1} + {\bf D}^{-1}{\bf V}^\top({\bf S}^{-1}
+ {\bf V}{\bf D}^{-1}{\bf V}^\top)^{-1}{\bf V}{\bf D}^{-1}$. That
leads to a low-rank version of Algorithm~\ref{alg:stochasticbound},
which requires only $\mathcal{O}(knd)$ work per iteration which is
pseudo-linear in dimension if $k$ is assumed to be a logarithmic or
constant function of $d$.

\section{A motivating example}
\label{sec:Motivatingexample}
\begin{figure}[h]
  \center
\setlength\tabcolsep{0pt}
\begin{tabular}{cc}
\includegraphics[width = 1.3in, height = 1.3in]{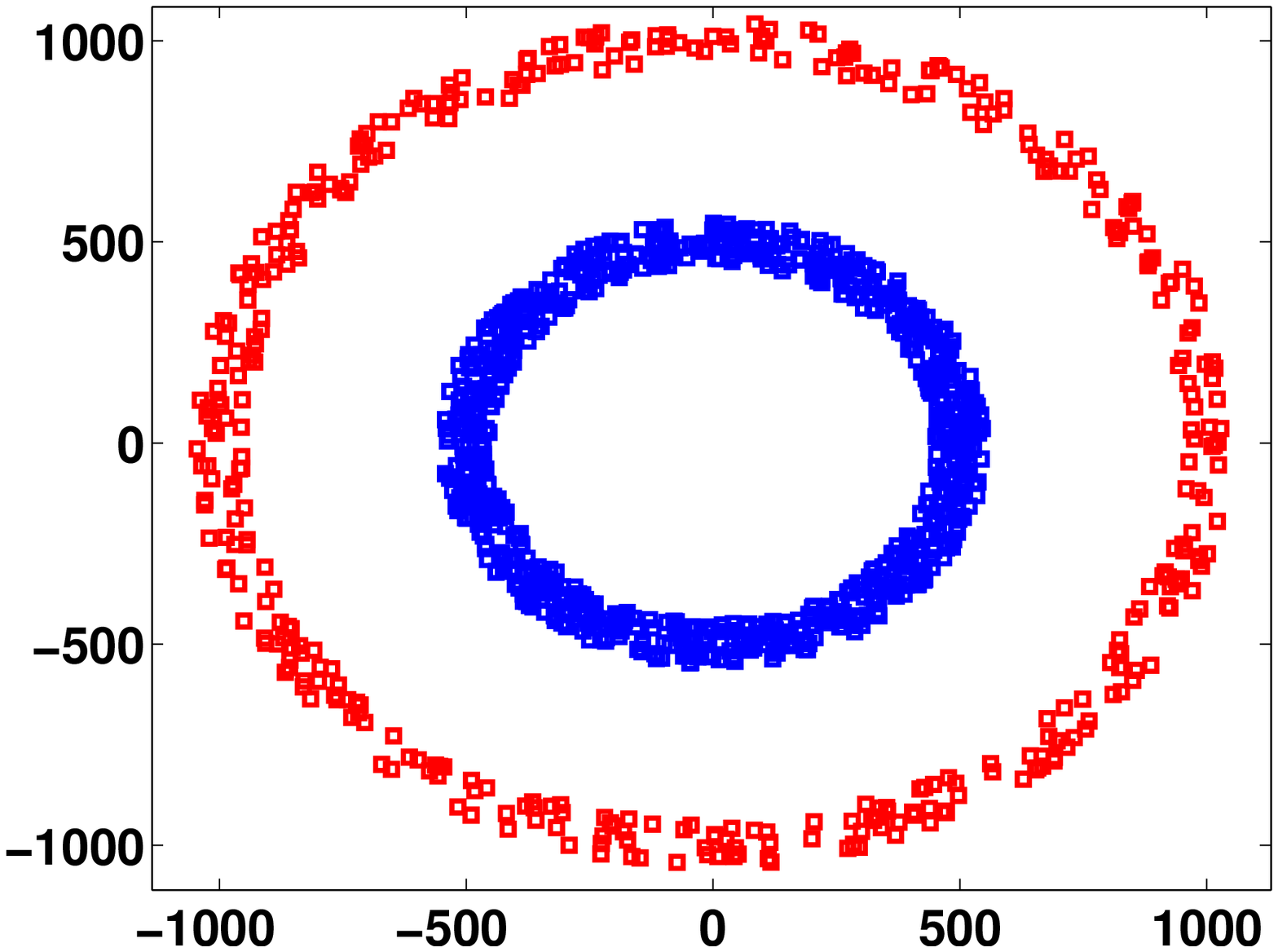}
\includegraphics[width = 1.3in, height = 1.3in]{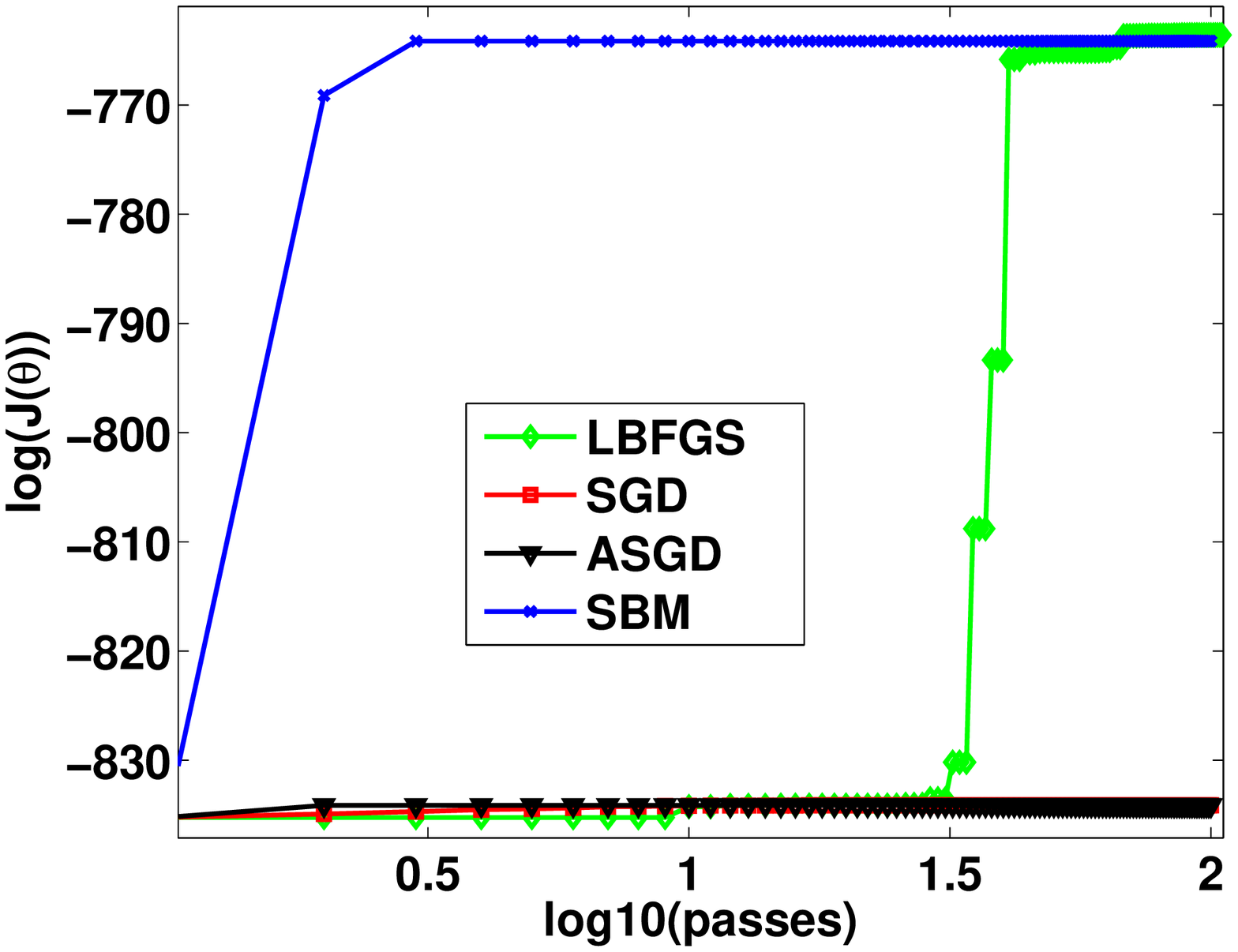}
\includegraphics[width = 1.3in, height = 1.3in]{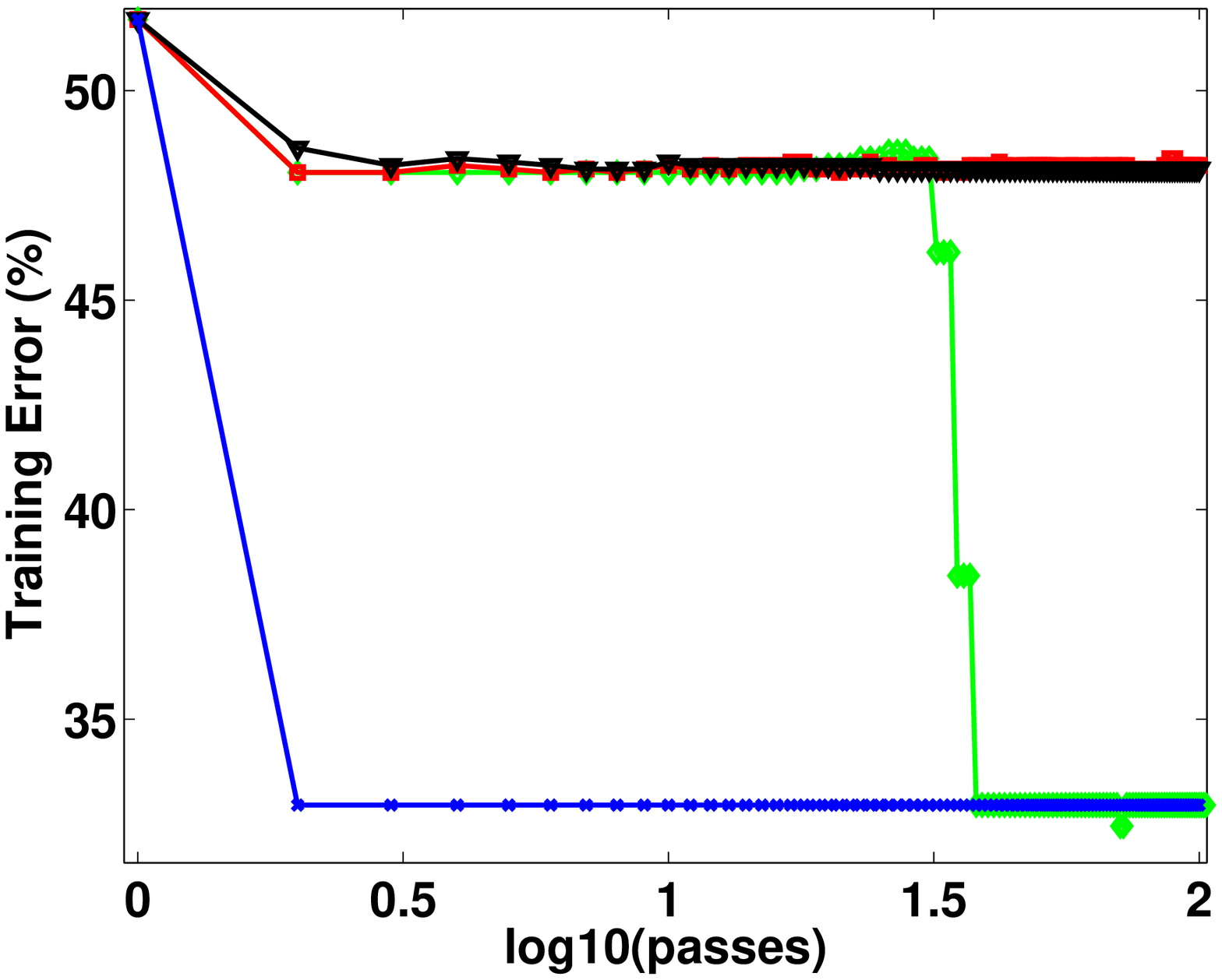} 
\includegraphics[width = 1.3in, height = 1.3in]{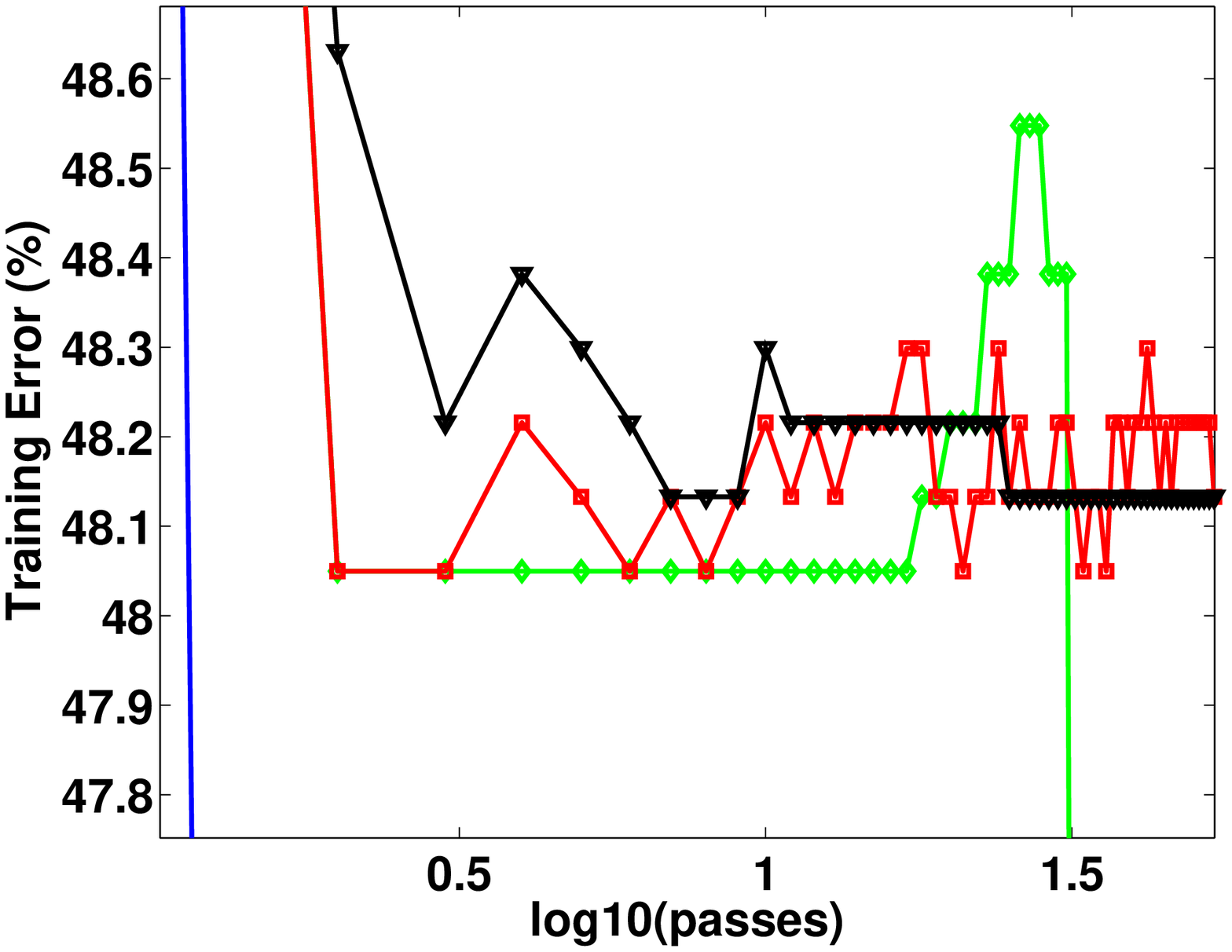}\\
\includegraphics[width = 1.3in, height = 1.3in]{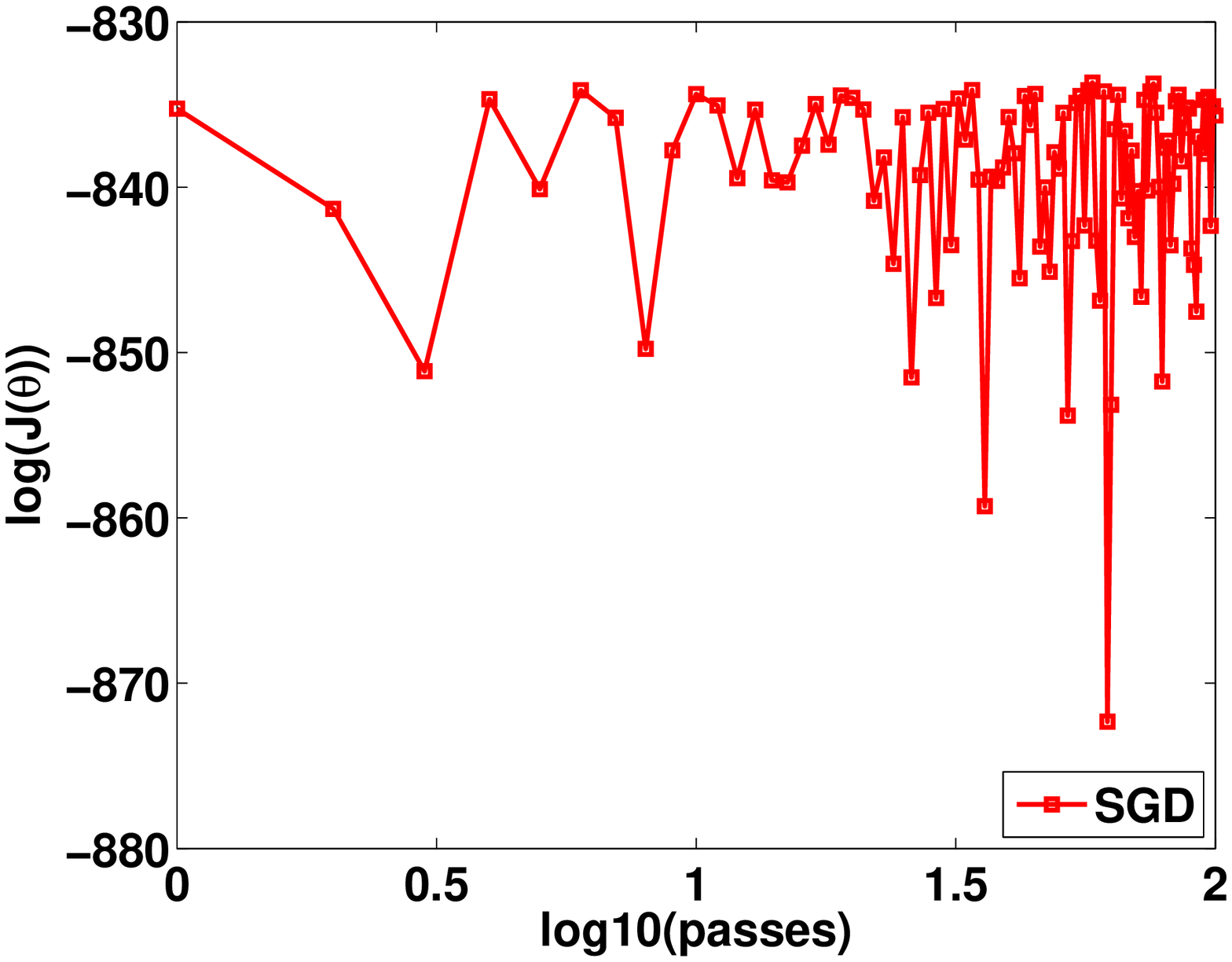}
\includegraphics[width = 1.3in, height = 1.3in]{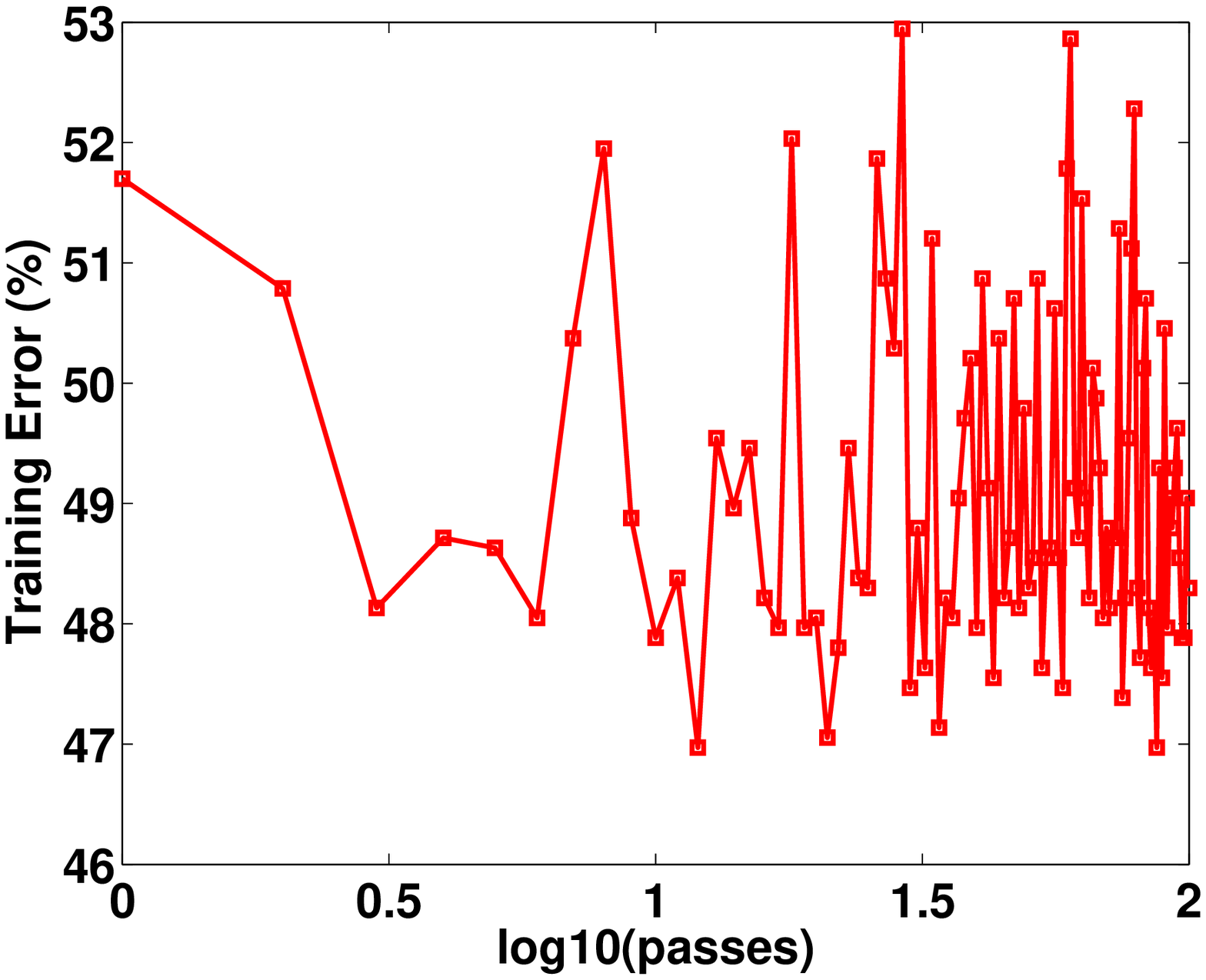}
\includegraphics[width = 1.3in, height = 1.3in]{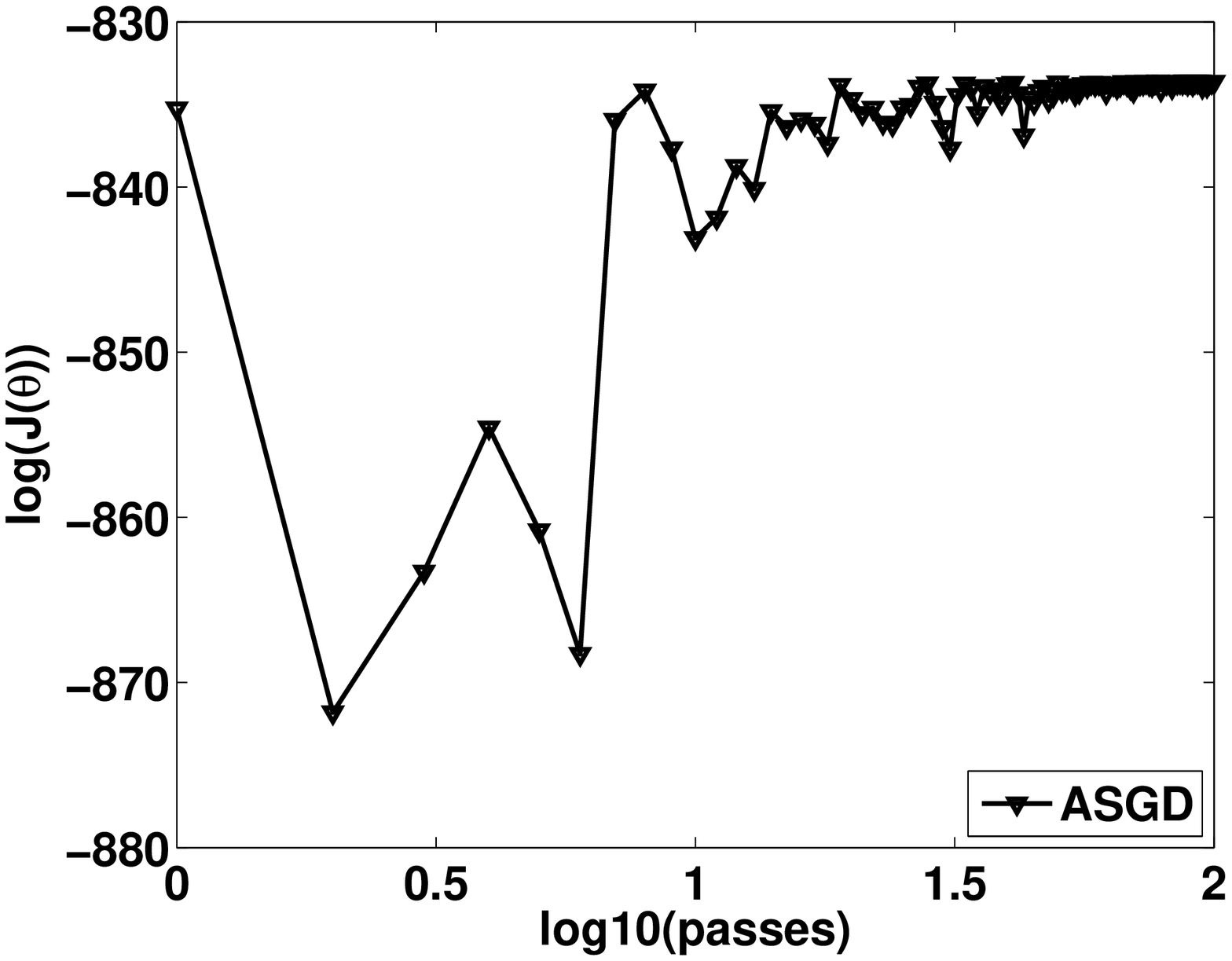} 
\includegraphics[width = 1.3in, height = 1.3in]{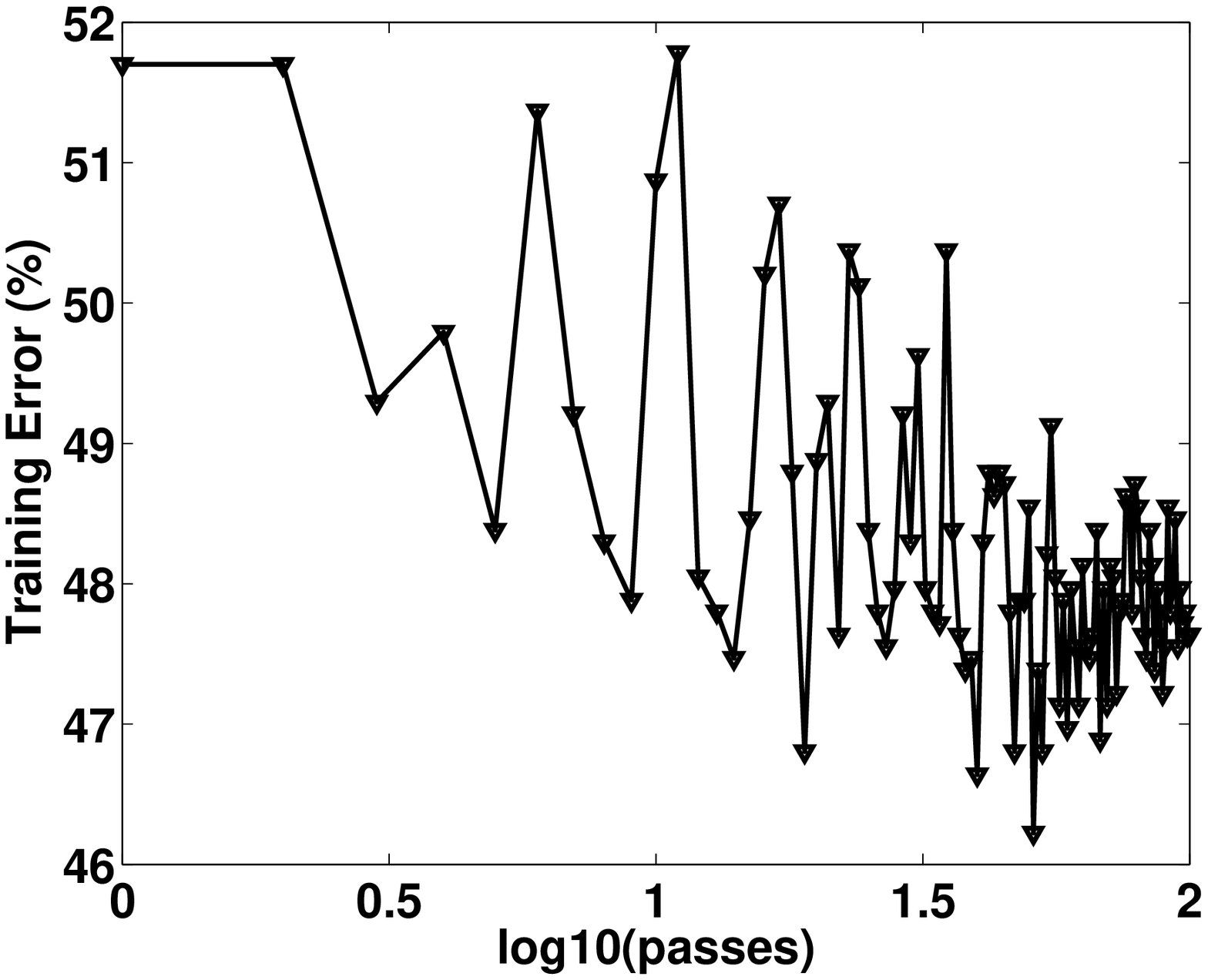}
\end{tabular}
\caption{A comparison of LBFGS, SGD, ASGD and SBM for $l_2$-regularized logistic regression. \textbf{From left to right, first row:} the data-set, training log-likelihood and error (original and zoomed) vs. passes through the data for LBFGS, SGD ($\eta_0 = 10^{-8}$, $m = 1$), ASGD ($\eta_0 = 10^{-5}$, $\tau = 1$, $m = 1$) and SBM ($\eta_0 = \frac{1}{t}$, $m = 1$), \textbf{second row}: SGD training log-likelihood and error ($\eta_0 = 10^{-7}$) and ASGD training log-likelihood and error ($\eta_0 = 10^{-4}$, $\tau = 1$) vs. passes through the data. $\lambda = 10^{1}$.}
\label{fig:toydataset}
\end{figure}

Consider Figure~\ref{fig:toydataset} which is an example of a binary classification problem which exposes some of difficulties with SGD. Intuitively, in this example, the gradients in the SGD update rule will point in almost random directions which could lead to very slow progress. Four training algorithms will be compared: SGD, ASGD, stochastic bound majorization algorithm (SBM) and LBFGS. We have tried several parameter settings for SGD and ASGD. The range of tested step size $\eta_0$ was as broad as $[10^{-12},1]$. The SGD method however exhibits high instability until the step size is reduced to an unreasonably small value such as $10^{-8}$ (for comparison we also show SGD performance for $\eta_0 = 10^{-7}$). The reason for is that this is a highly symmetric and non-linearly separable data-set. Therefore, the information captured in the gradients is extremely noisy which causes SGD to oscillate and fail to converge in practice. For ASGD we obtained the best and stable result for $\eta_0 = 10^{-5}$ (for comparison we also show ASGD performance for $\eta_0 = 10^{-4}$) and $\tau = 1$ (larger values of $\tau$ weaken performance). For both methods we tested the mini-batch size $m$ from  $1$ (a single data point) to $10$ and noticed no meaningful difference. Clearly both SGD and ASGD are stuck in solutions that are only slightly better than random guessing. Meanwhile SBM (with $m = 1$ and a constant step size $\eta_0 = \frac{1}{t}$) finds the same solution as a batch LBFGS method. It does so with a single pass through the data and simultaneously outperforming the competitor methods. We emphasize that all methods used as comparators to our algorithm were well-tuned, i.e. the initial gain $\eta_0$ is set as high as
possible for each method while maintaining stability.

\begin{figure}[h]
\vspace{-0.1in}
  \center
\setlength\tabcolsep{0pt}
\begin{tabular}{cc}
\includegraphics[width = 1.3in, height = 1.3in]{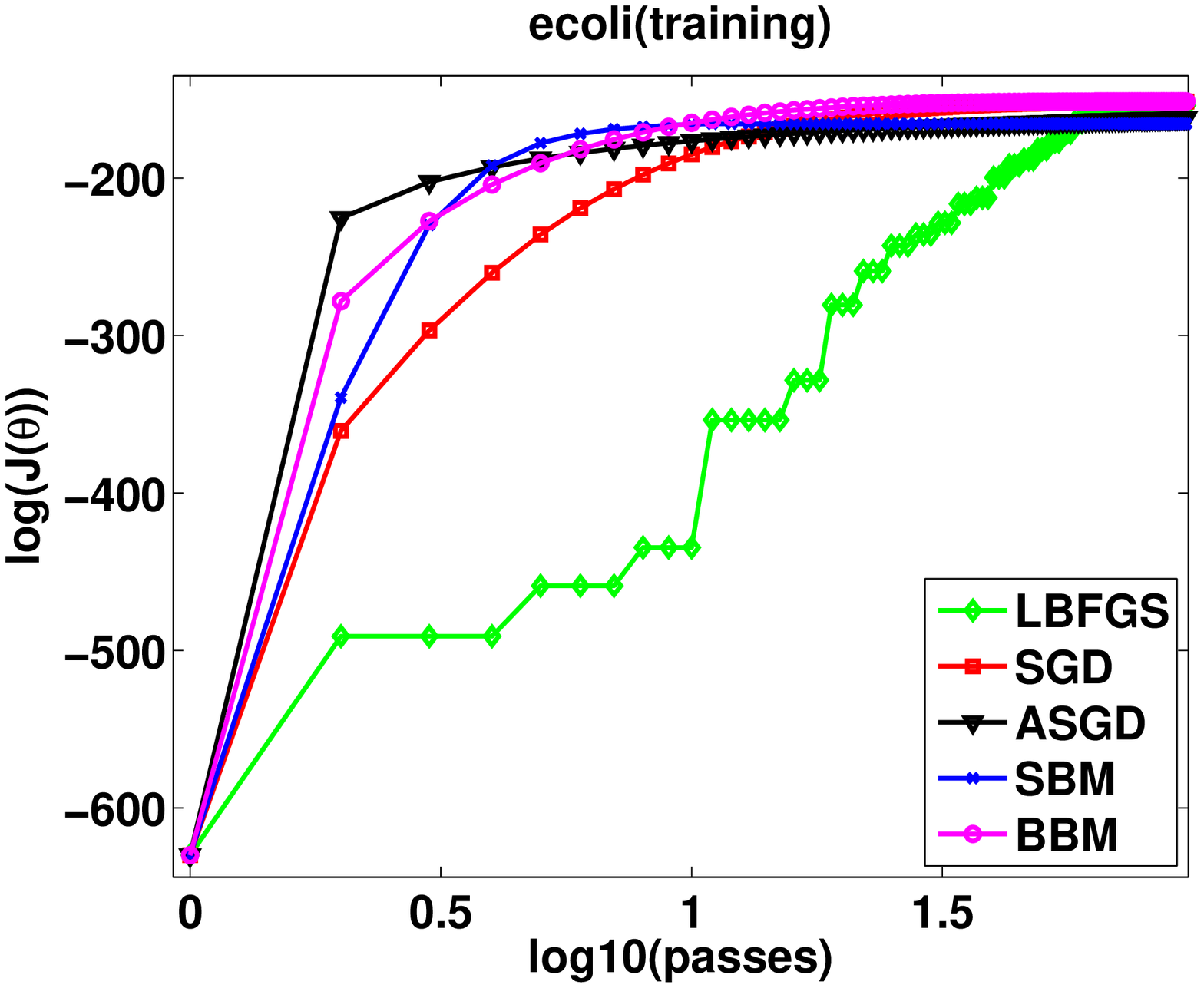}
\includegraphics[width = 1.3in, height = 1.3in]{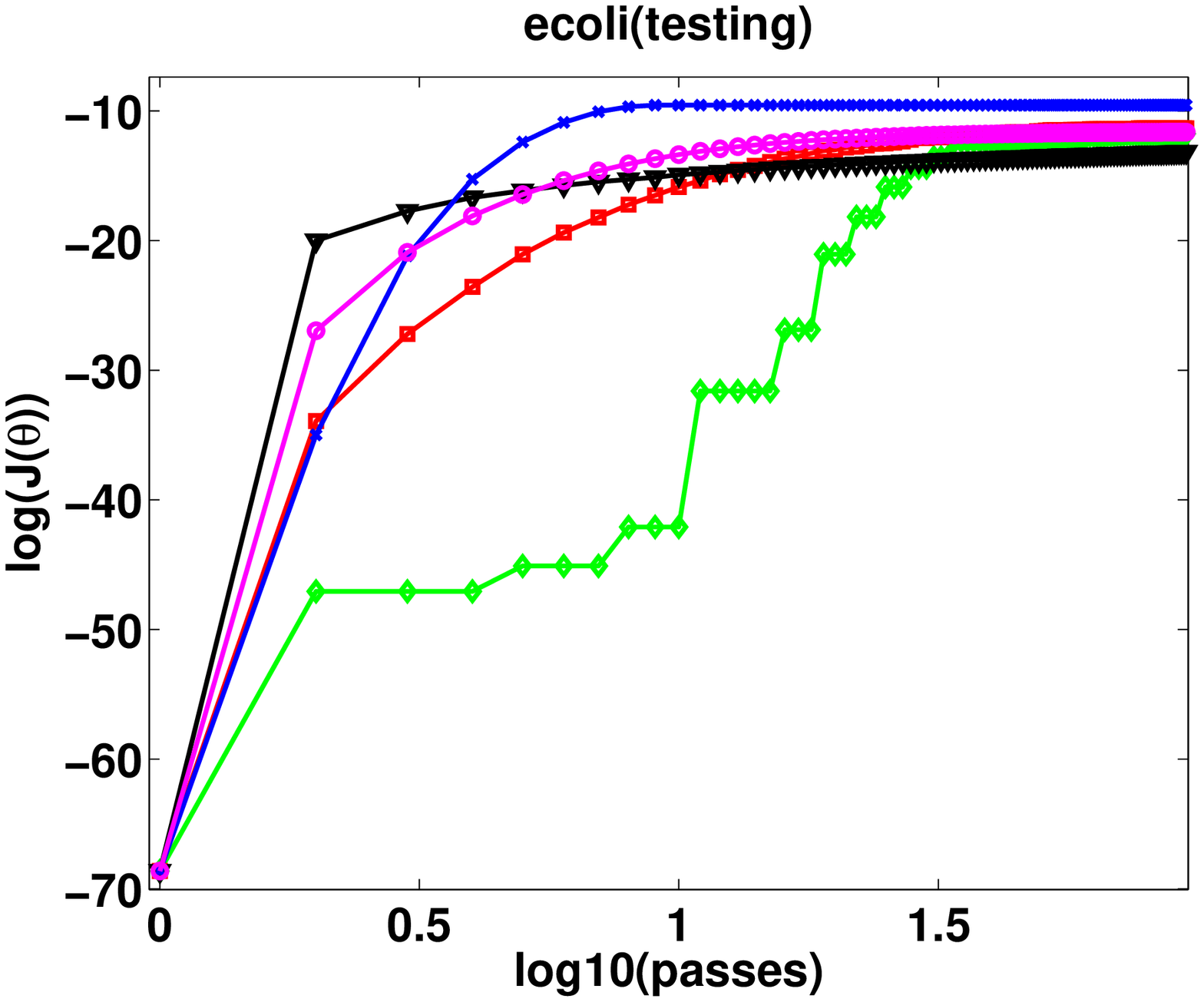}
\includegraphics[width = 1.3in, height = 1.3in]{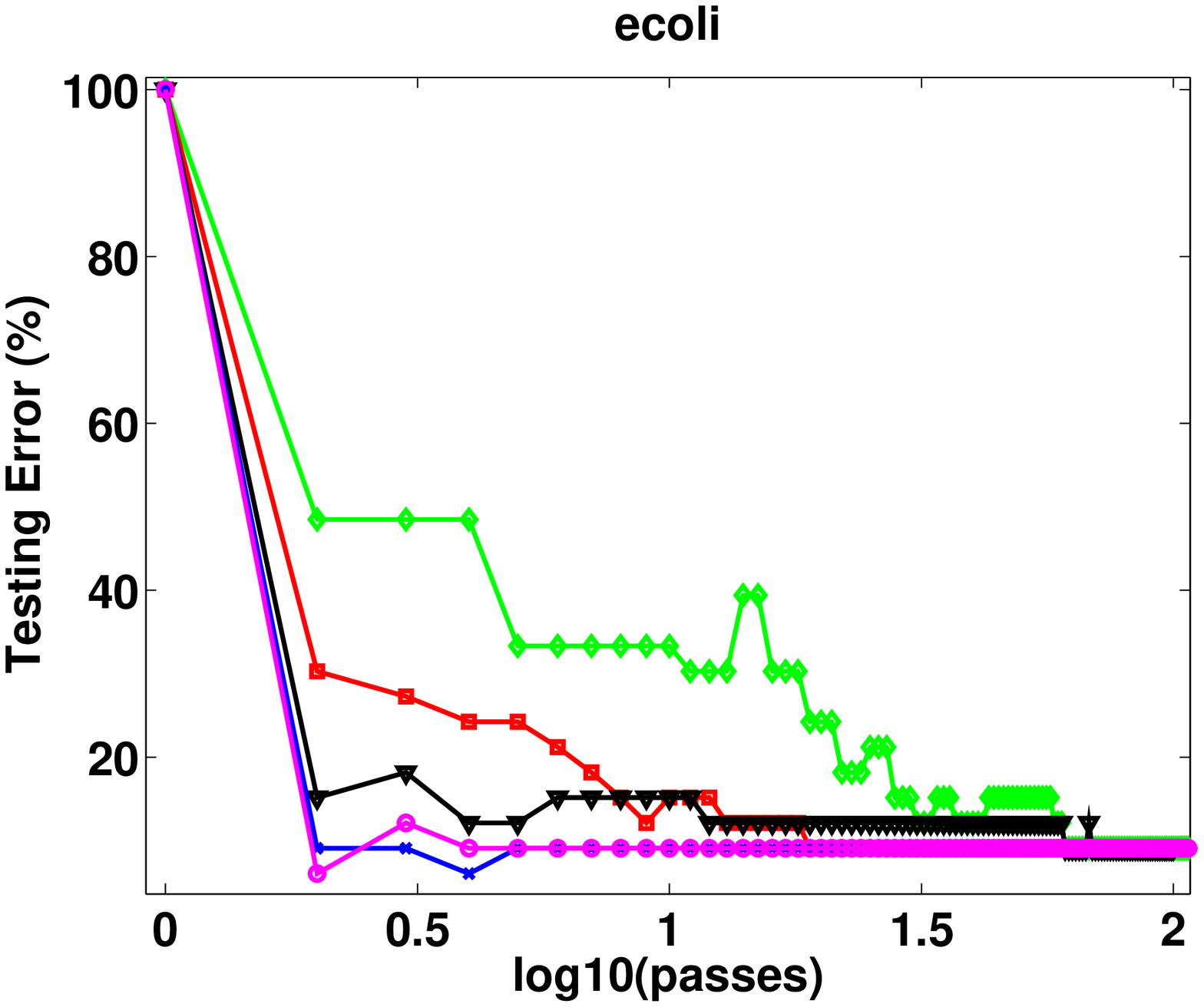} 
\includegraphics[width = 1.3in, height = 1.3in]{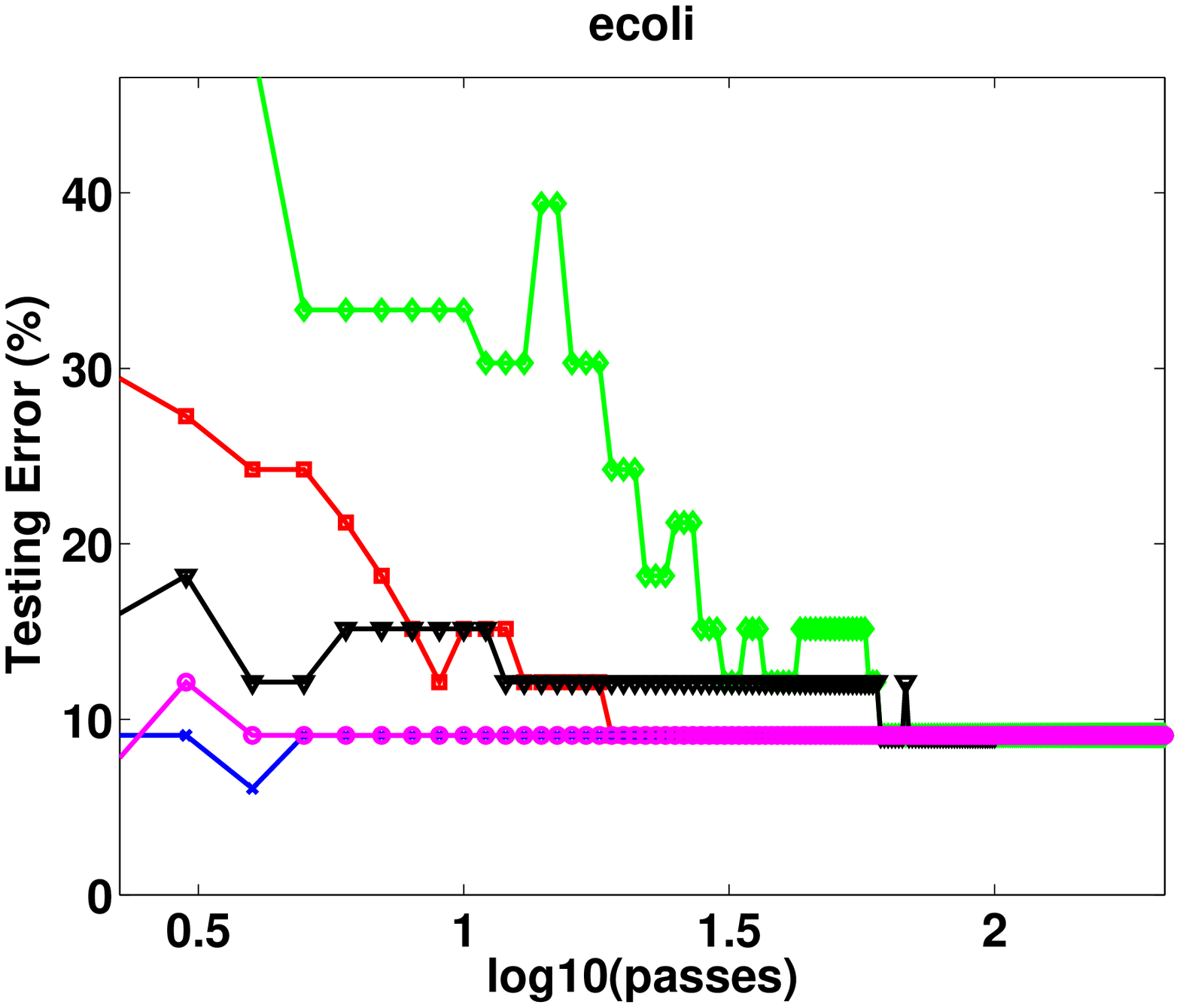}\\
\includegraphics[width = 1.3in, height = 1.3in]{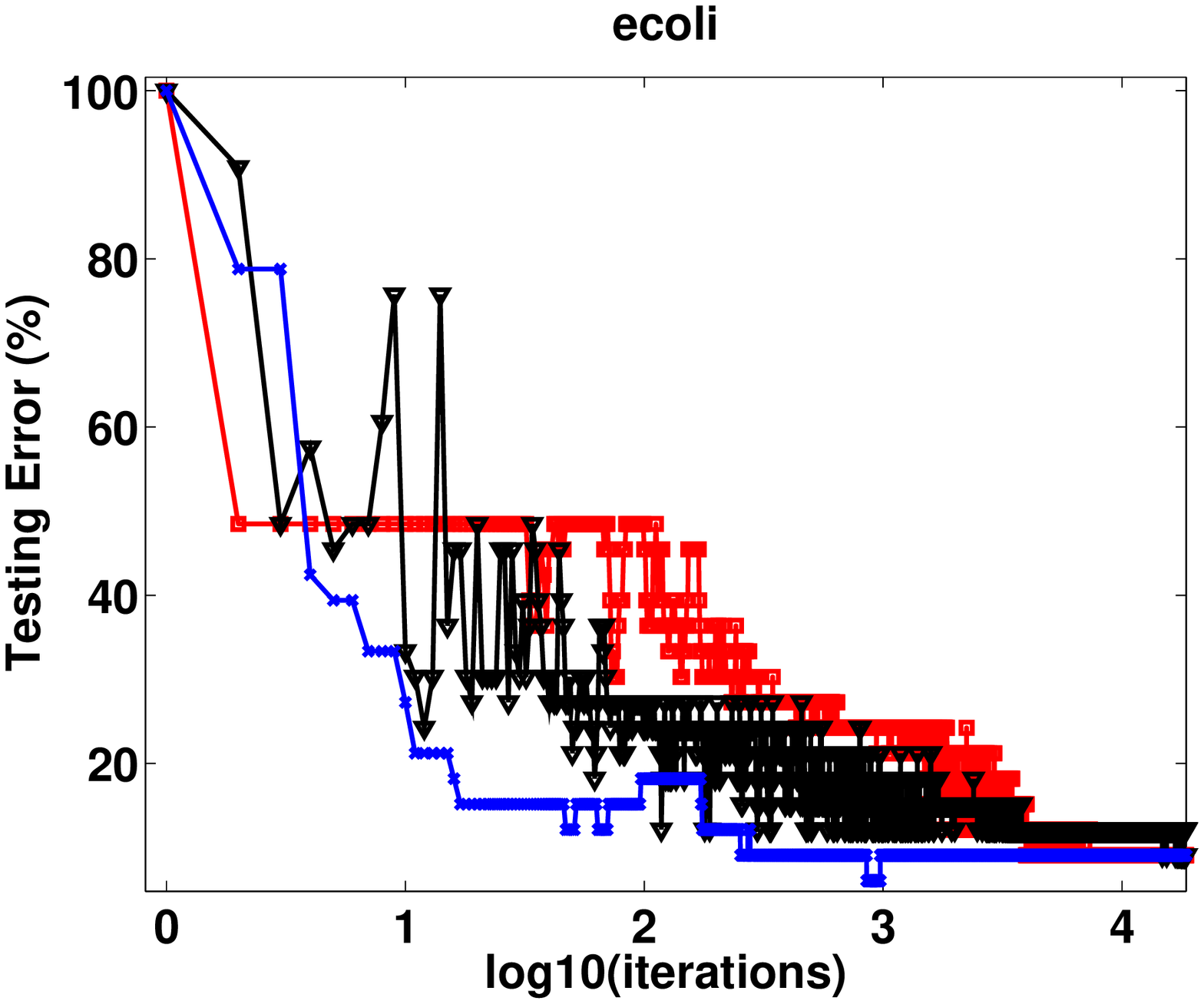} 
\includegraphics[width = 1.3in, height = 1.3in]{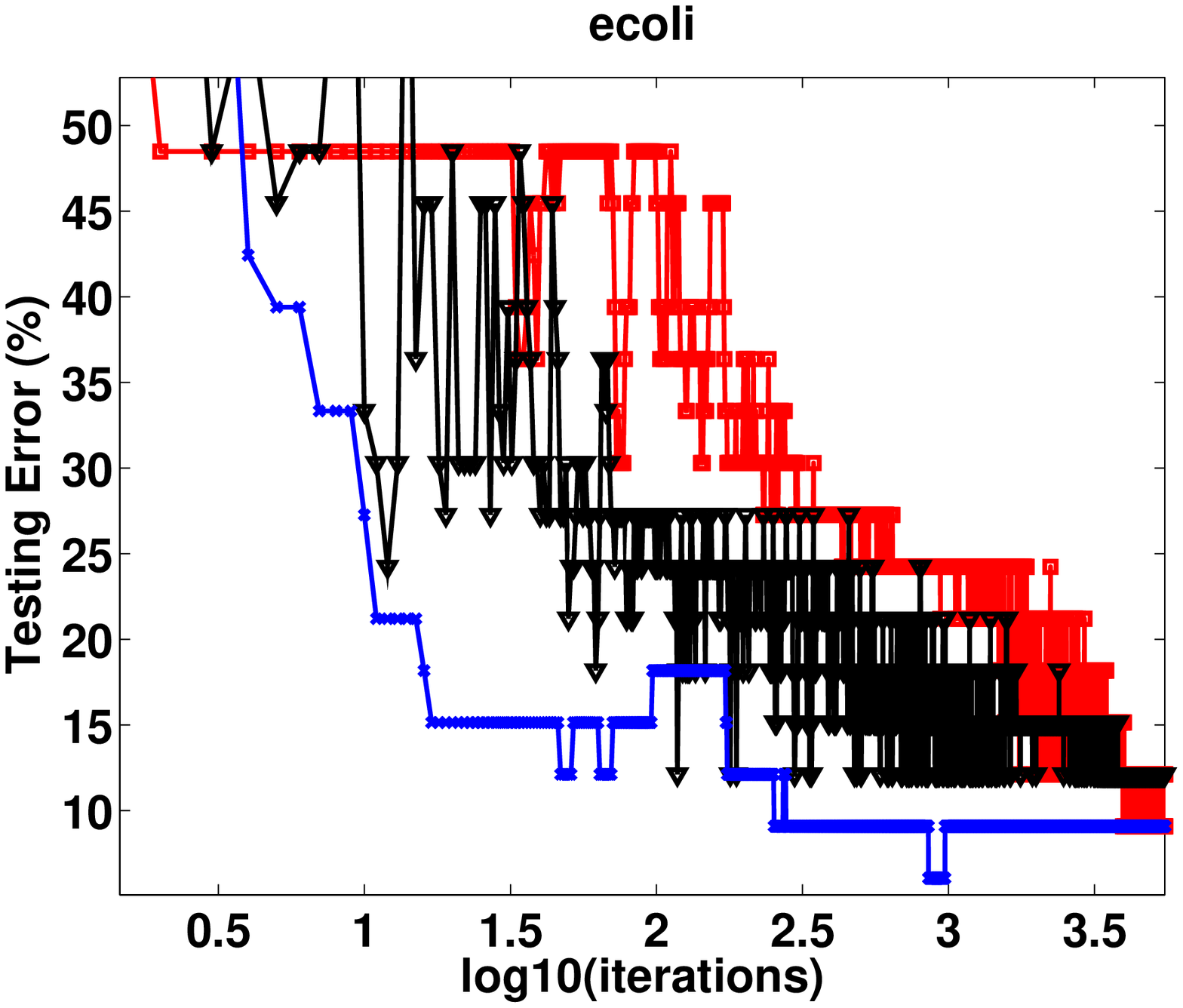}
\end{tabular}
\caption{A comparison of LBFGS, SGD, ASGD, SBM and BBM for $l_2$-regularized logistic regression. \textbf{From left to right, first row:} training and testing log-likelihood and testing error (original and zoomed) vs. passes through the data for LBFGS, SGD ($\eta_0 = 10^{-2}$, $m = 10$), ASGD ($\eta_0 = 10^{-1}$, $\tau = 500$, $m = 10$), SBM ($\eta_0 = \frac{1}{t}$, $m = 1$) and BBM on the ecoli data-set, \textbf{second row}: testing error (original and zoomed) vs. iterations for LBFGS, SGD and ASGD on the same experiment.}
\label{fig:ecoli}
\end{figure} 

Next, we focus on the ecoli UCI data-set (\textsf{http://archive.ics.uci.edu/ml/}), a simple small-scale classification problem. We compare LBFGS, batch bound majorization method (BBM), SGD, ASGD and SBM. Results are summarized in Figure~\ref{fig:ecoli}. BBM, which was already shown to outperform leading batch methods \cite{JebCho12}, performs comparably to ASGD and SGD. The only method that beats BBM is SBM. In the first row of Figure~\ref{fig:ecoli}, we plot the objective with respect to the number of passes through the data. However, looking more closely at the objective with respect to each iteration (where a single iteration corresponds to a single update of the parameter vector), we note an interesting property of SBM: it clearly remains monotonic in its convergence despite its stochastic nature. This is in contrast to SGD and ASGD which (as expected) fluctuate much more noisily.

Note that, in all experiments in this section and the next, $90\%$ of the data is used for training and the rest for testing, the results are averaged over $10$ random initializations close to the origin and the regularization value $\lambda$ is chosen through crossvalidation. All methods were implemented in C++ using the mex environment under Matlab.

\section{Experiments}
\label{sec:Experiments}
\begin{figure}[h]
\vspace{-0.1in}
  \center
\setlength\tabcolsep{0pt}
\begin{tabular}{cc}
\includegraphics[width = 1.05in, height = 1.05in]{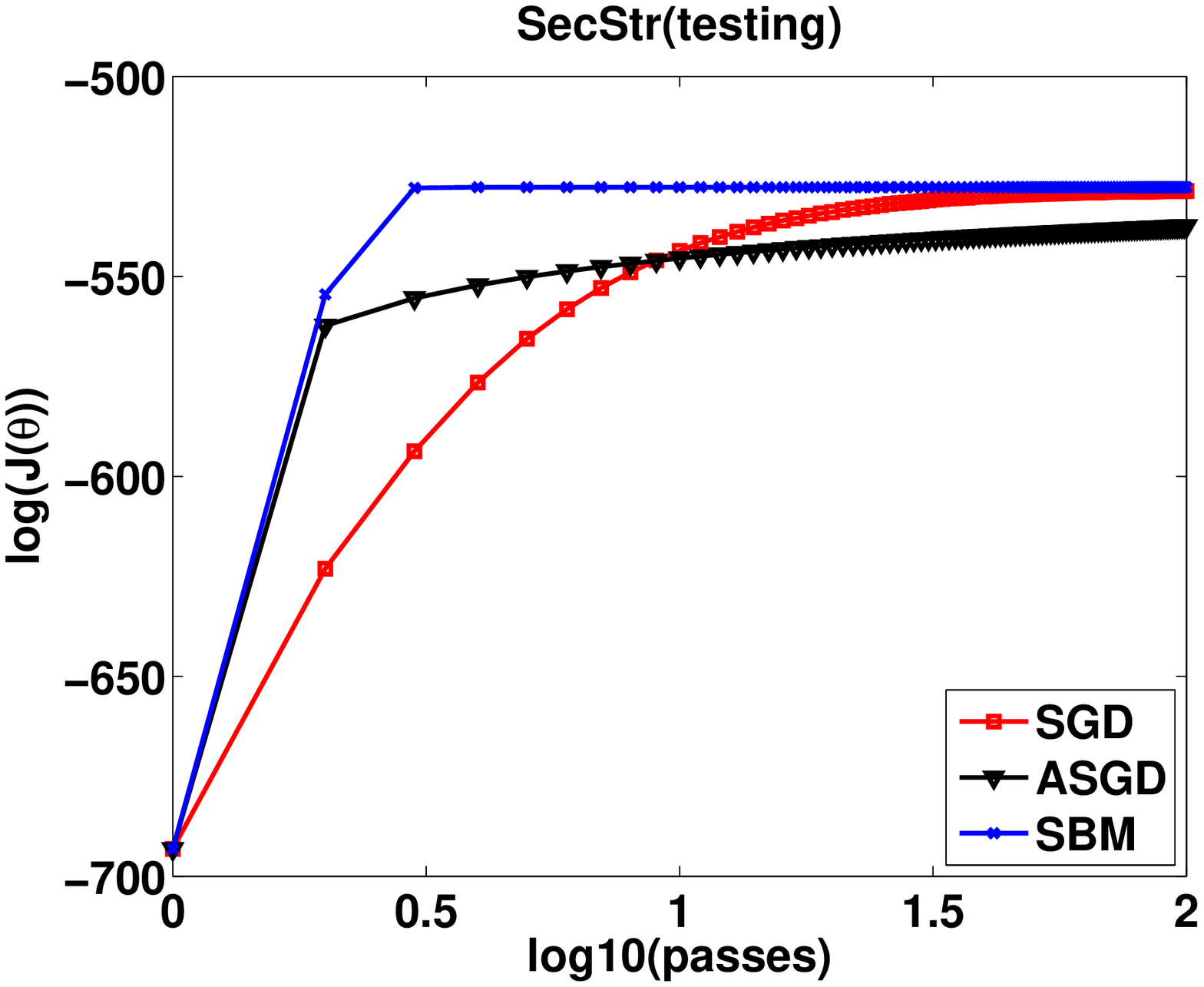}
\includegraphics[width = 1.05in, height = 1.05in]{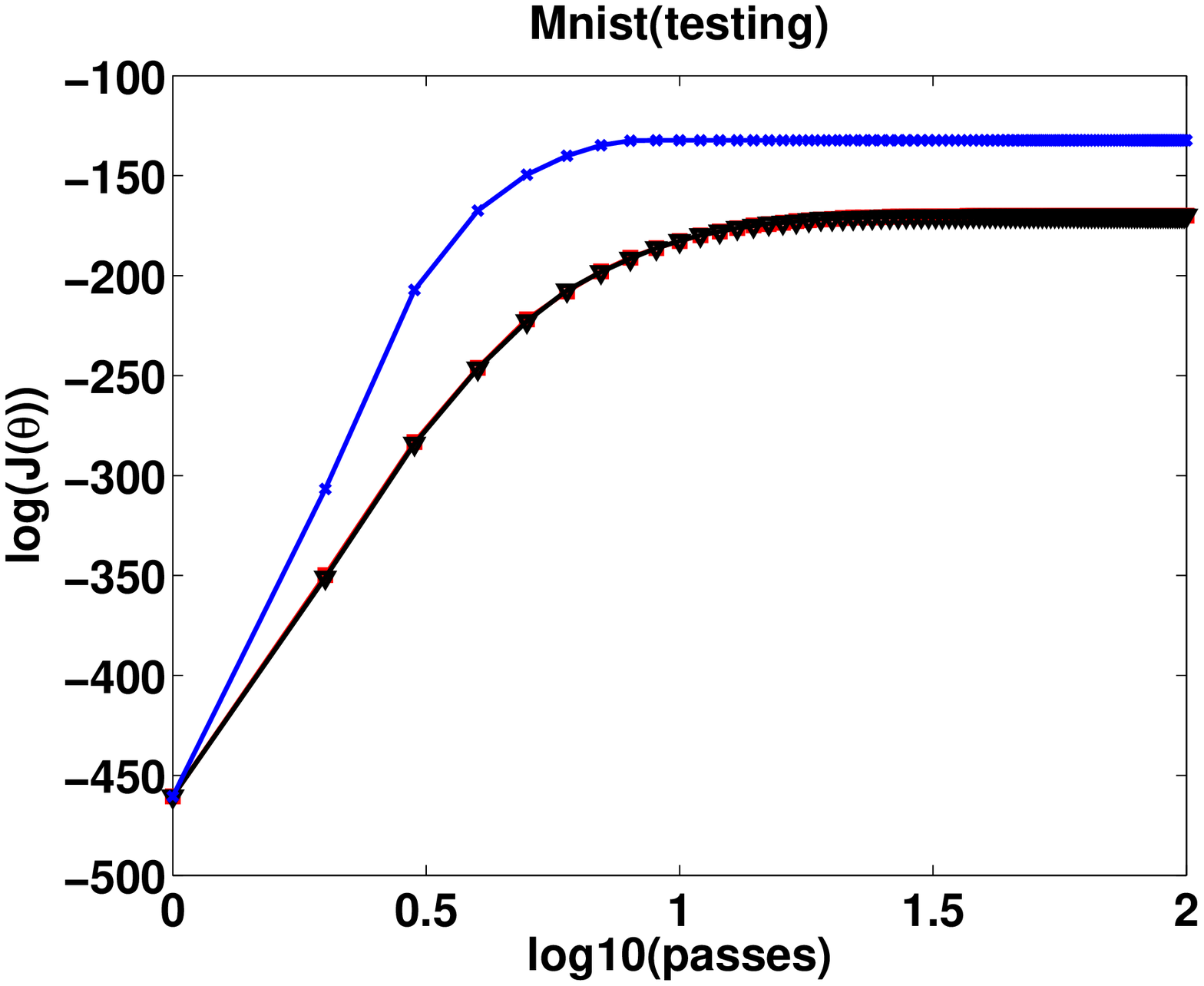}
\includegraphics[width = 1.05in, height = 1.05in]{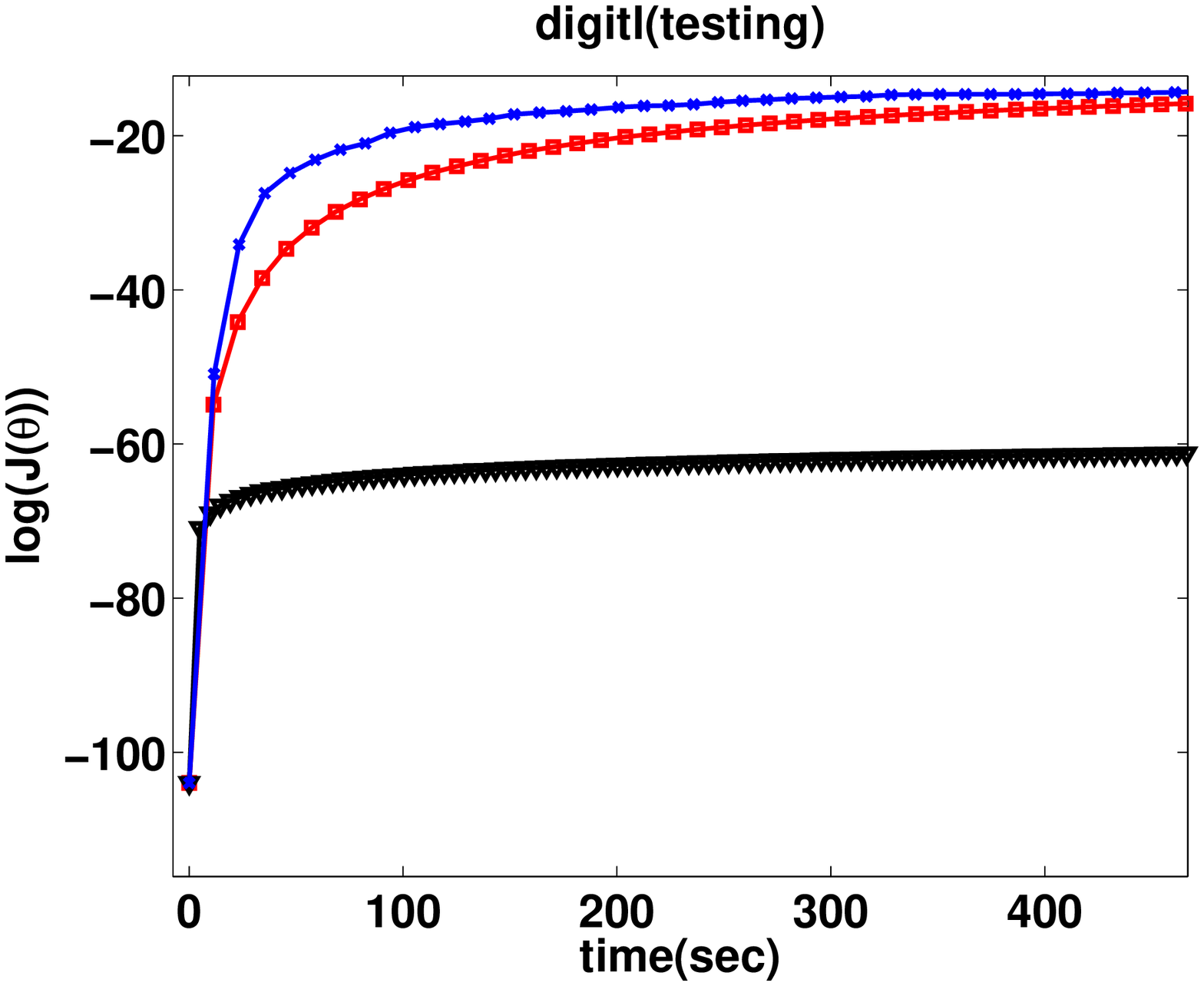}
\includegraphics[width = 1.05in, height = 1.05in]{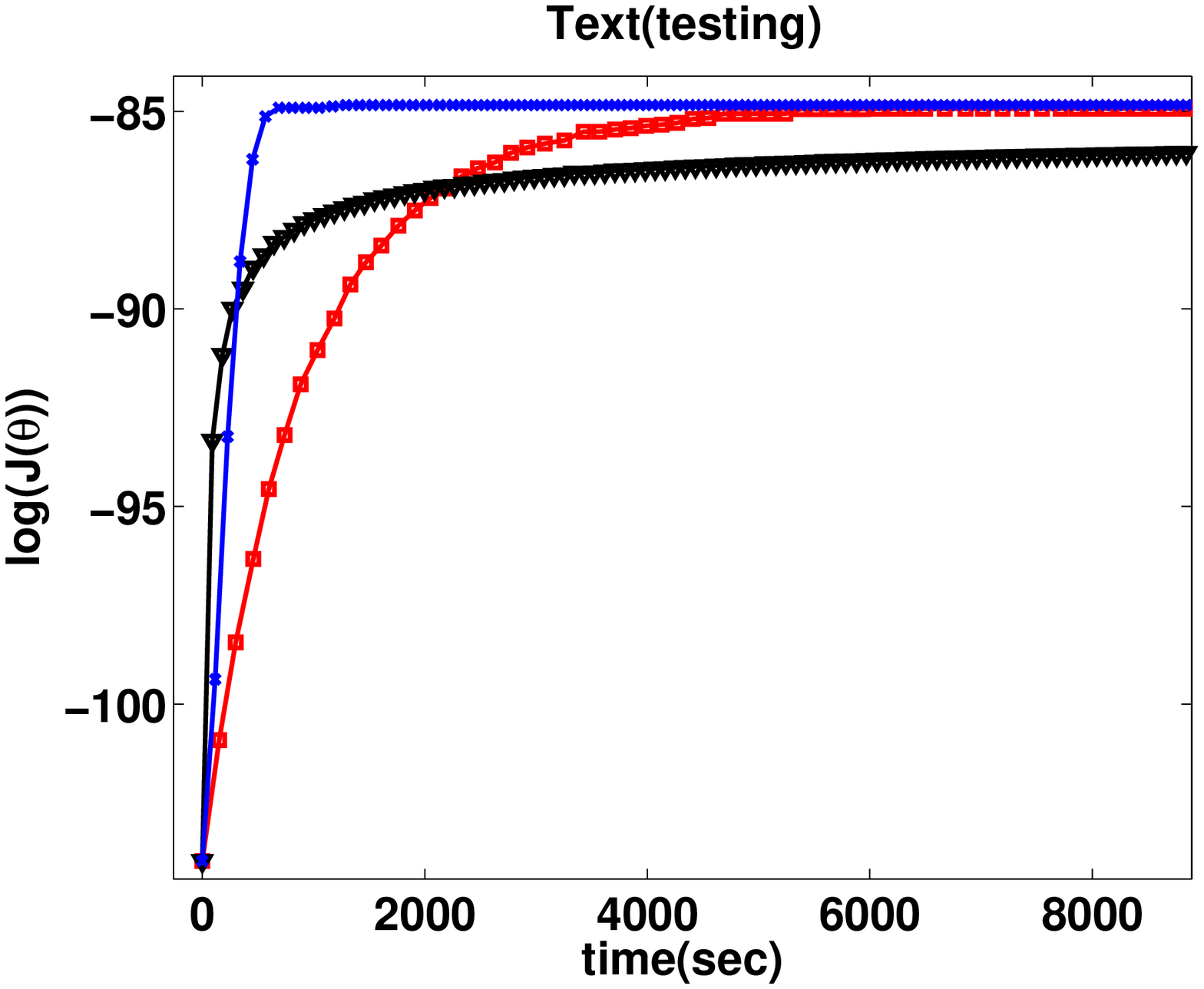}
\includegraphics[width = 1.05in, height = 1.05in]{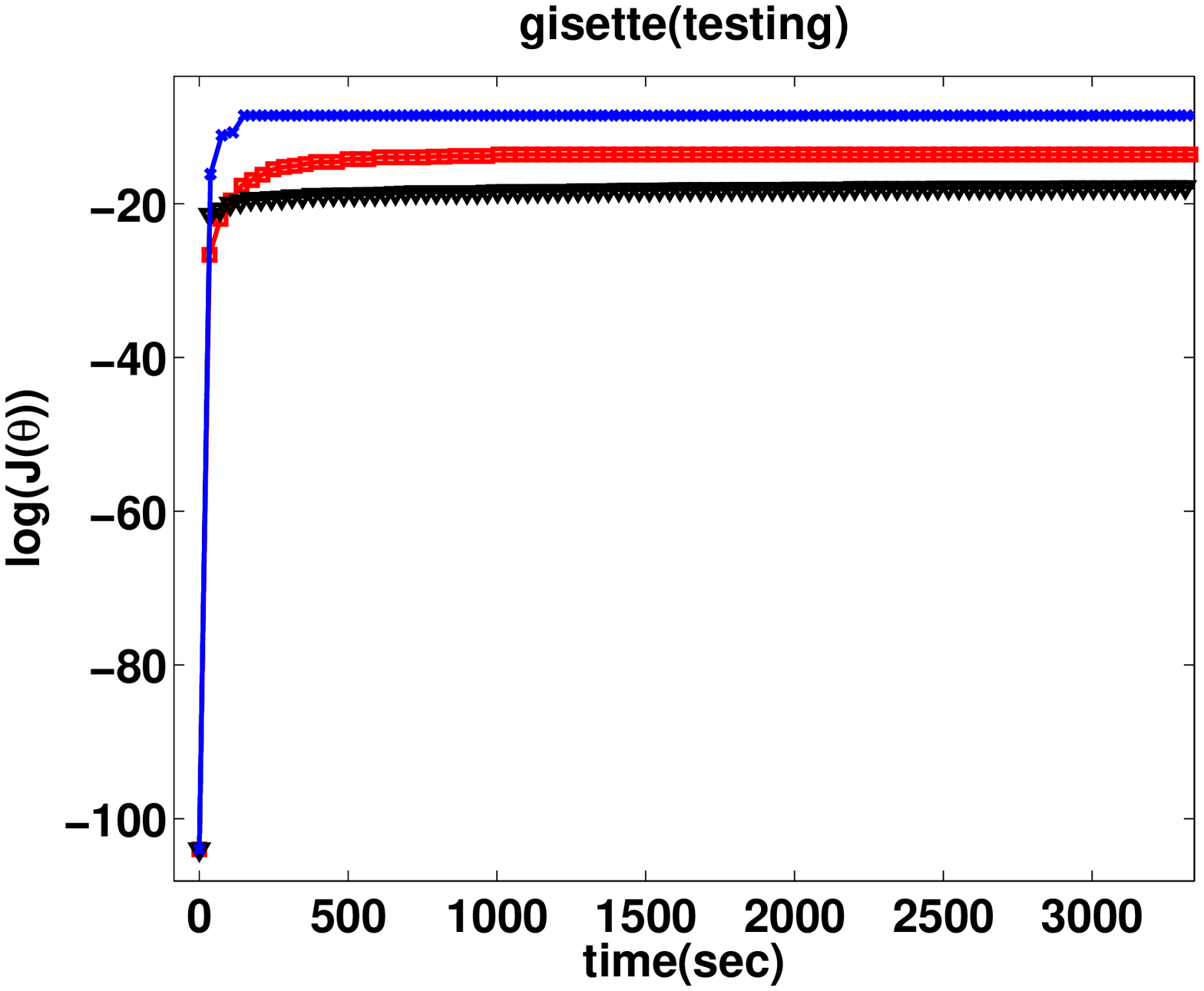}\\
\includegraphics[width = 1.05in, height = 1.05in]{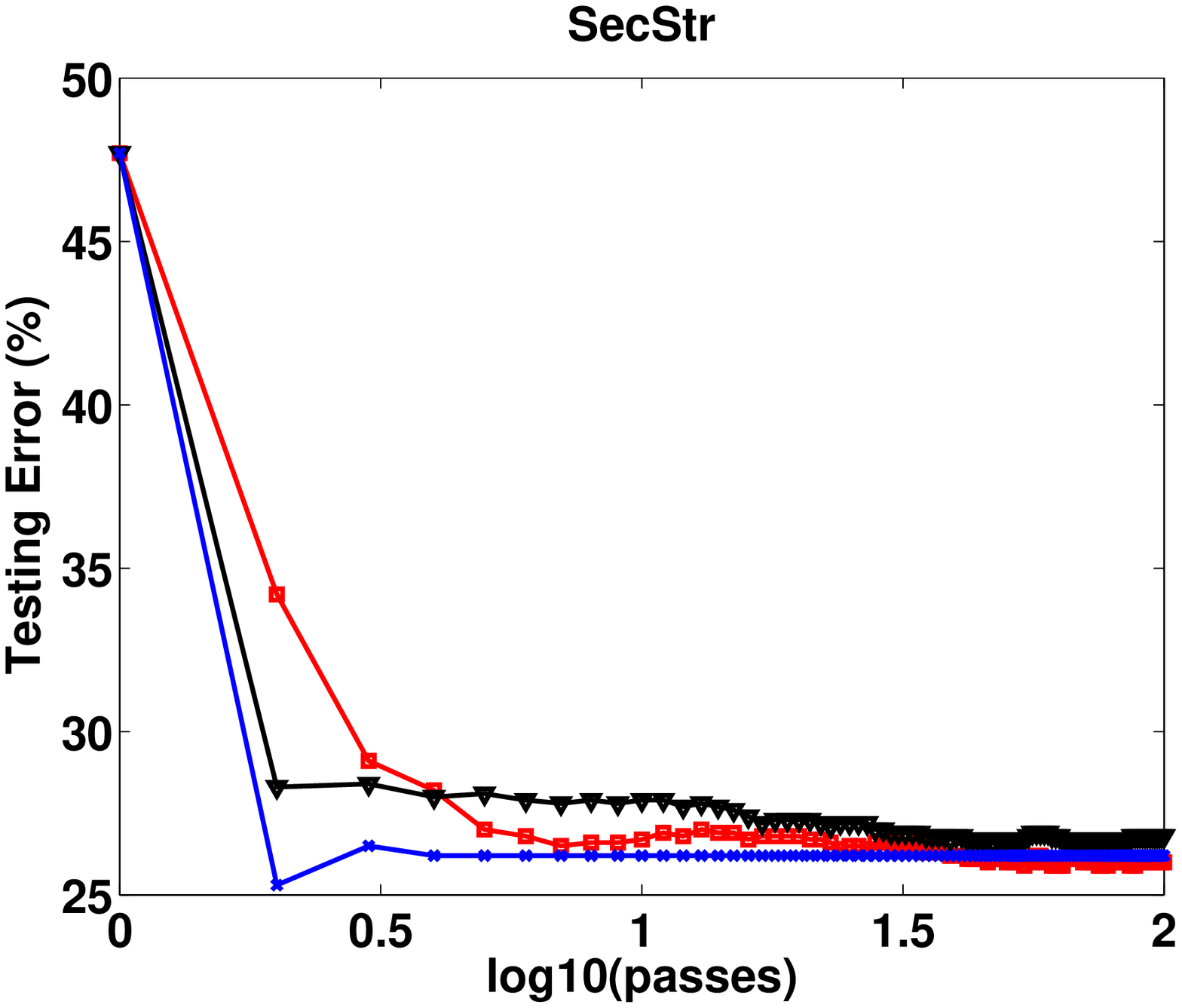}
\includegraphics[width = 1.05in, height = 1.05in]{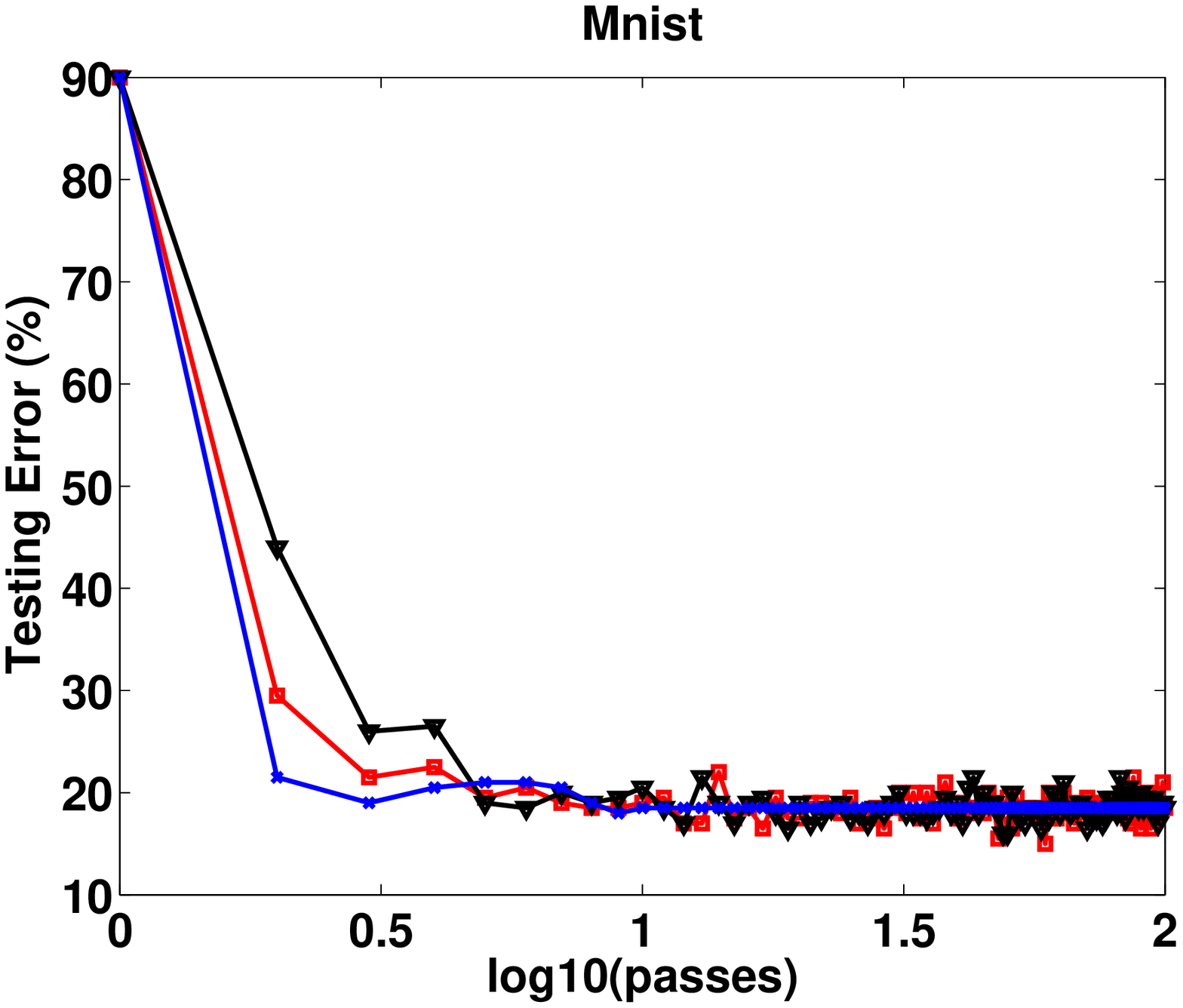}
\includegraphics[width = 1.05in, height = 1.05in]{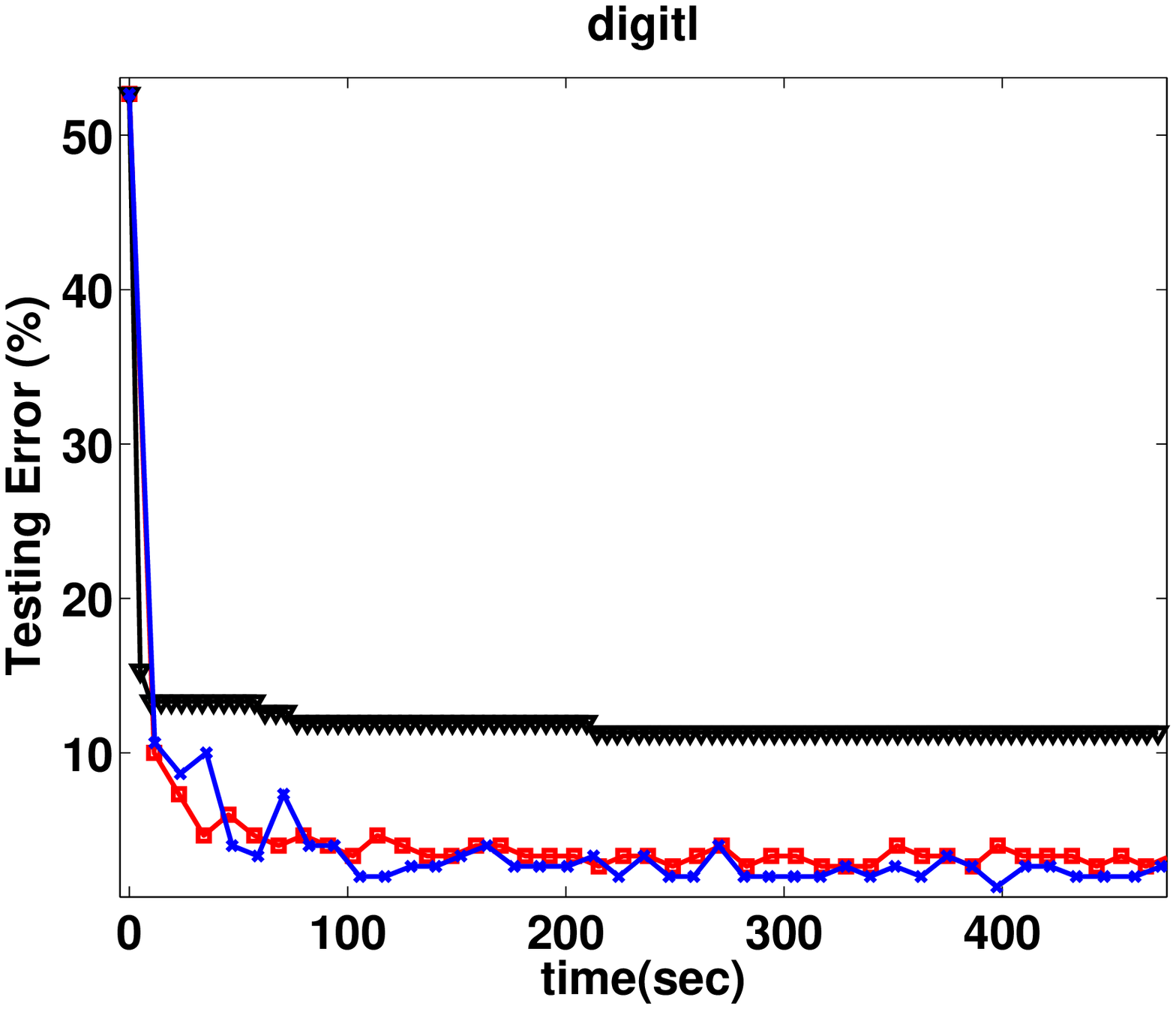} 
\includegraphics[width = 1.05in, height = 1.05in]{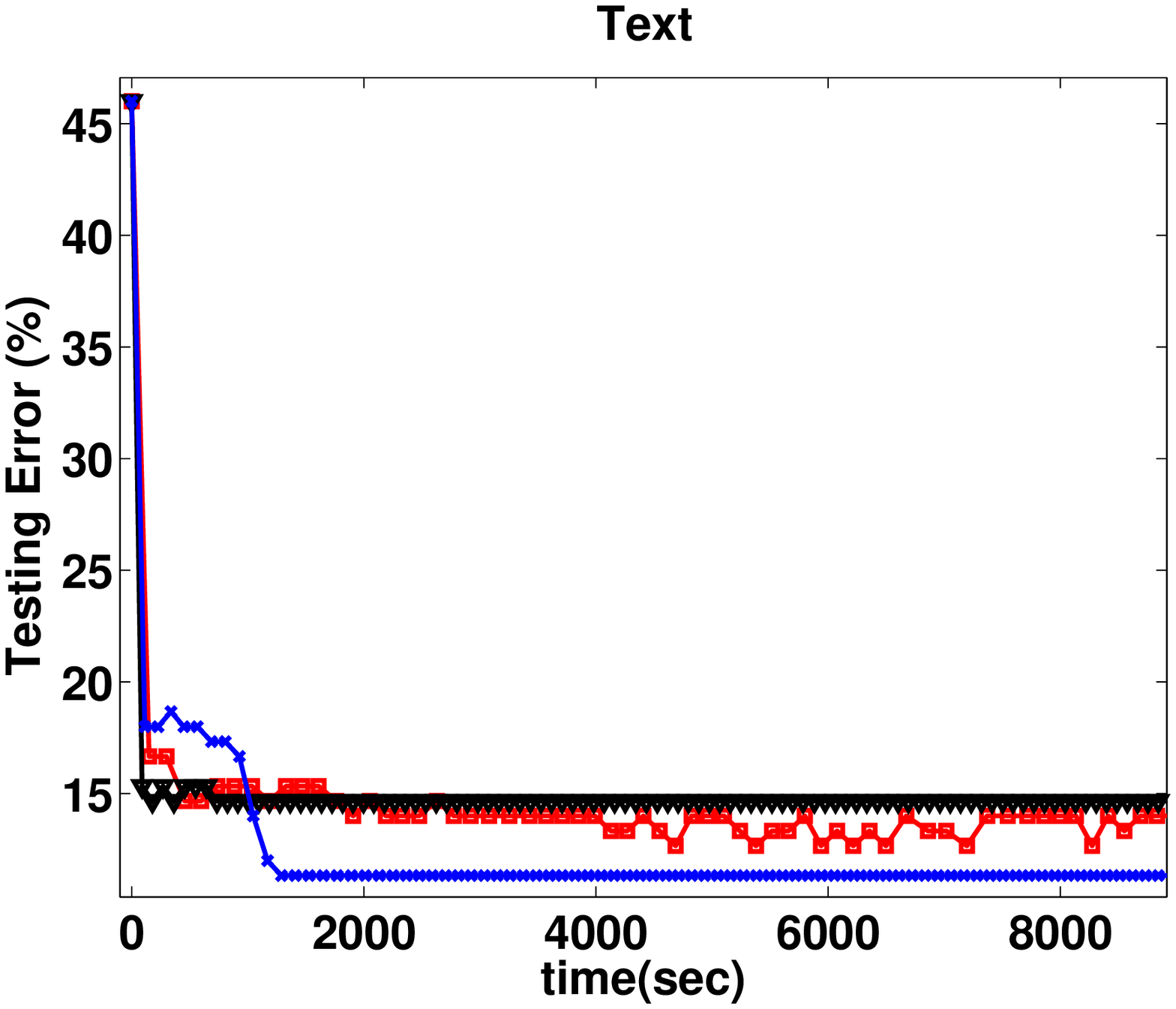}
\includegraphics[width = 1.05in, height = 1.05in]{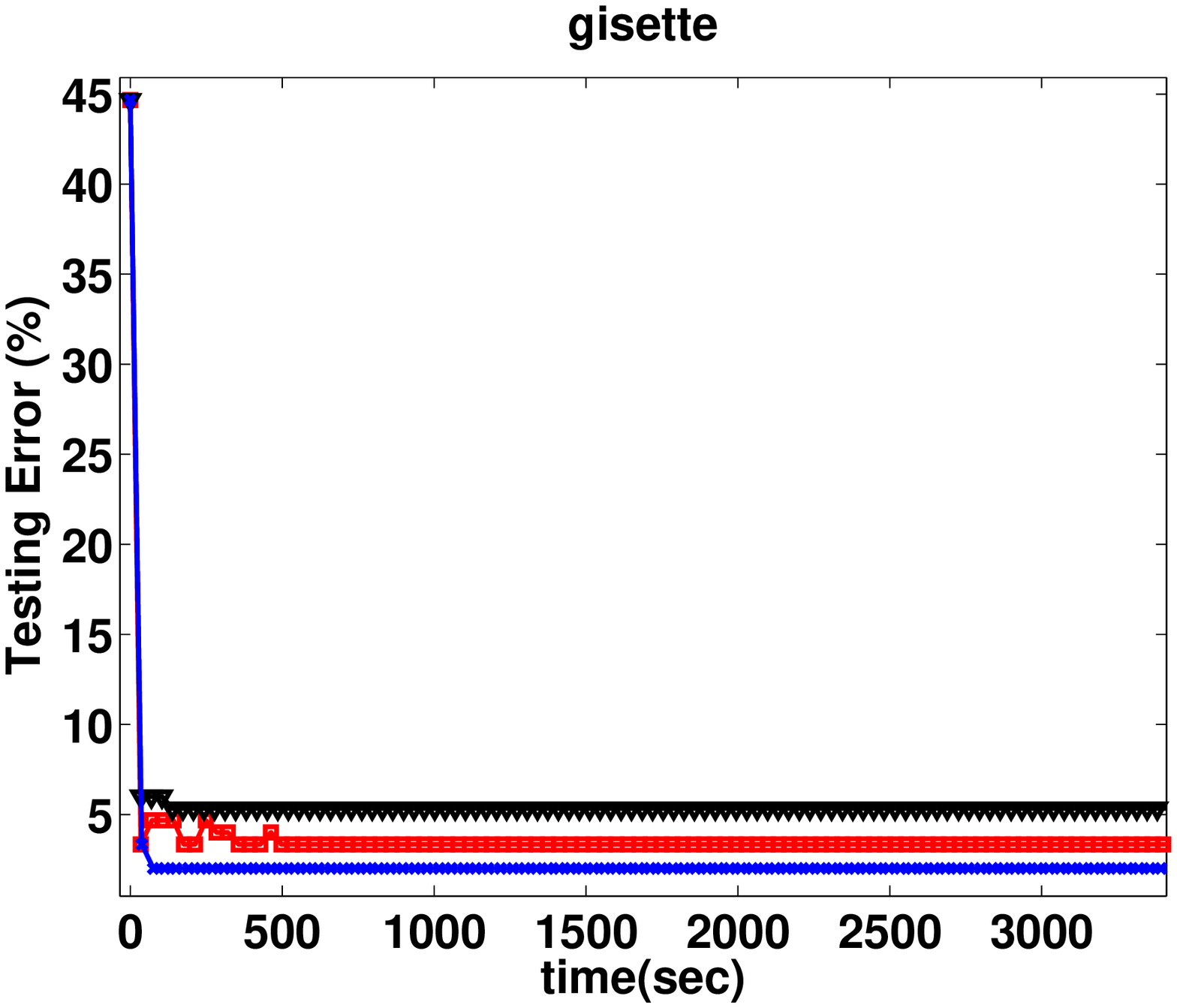}\\
\includegraphics[width = 1.05in, height = 1.05in]{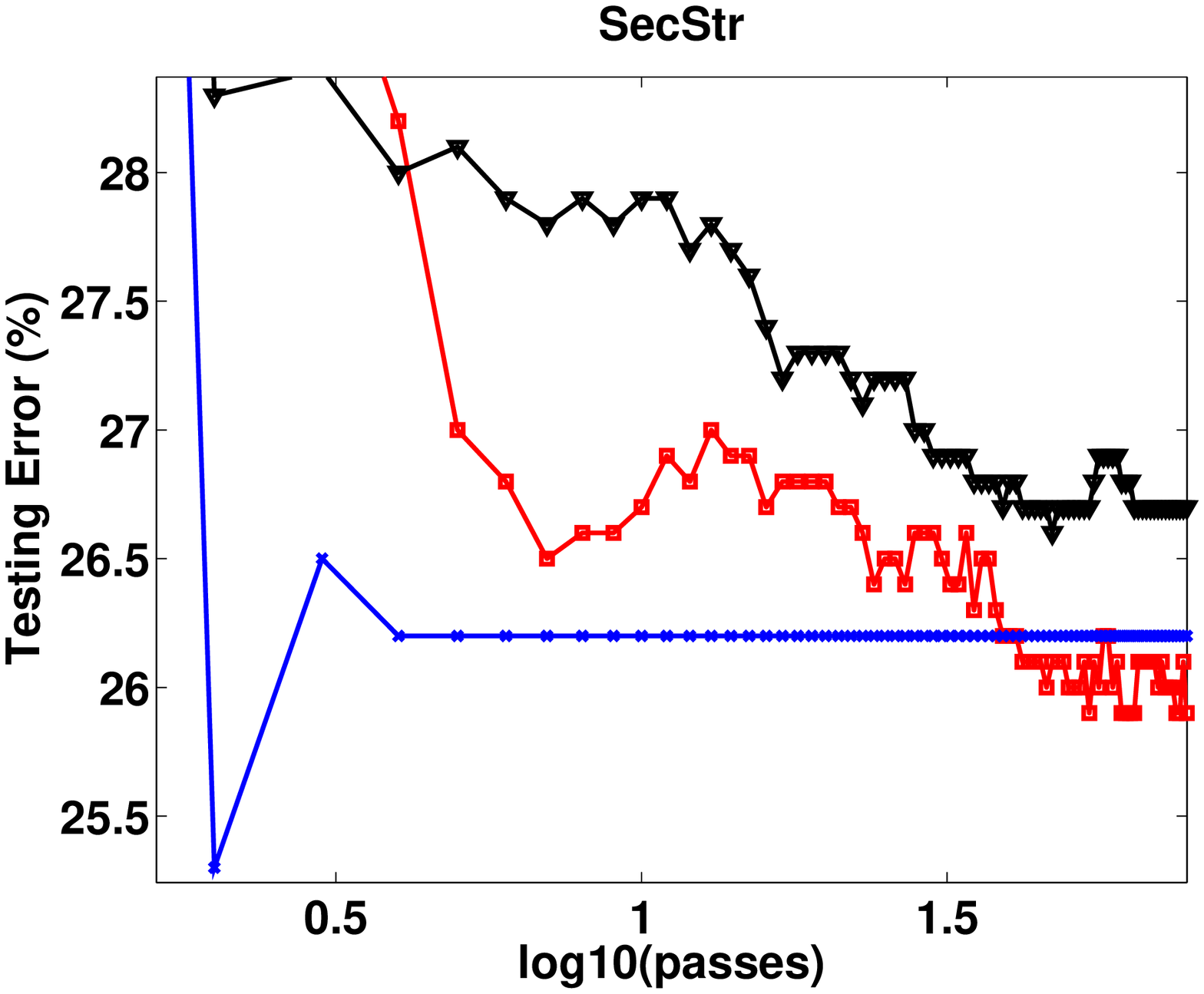}  
\includegraphics[width = 1.05in, height = 1.05in]{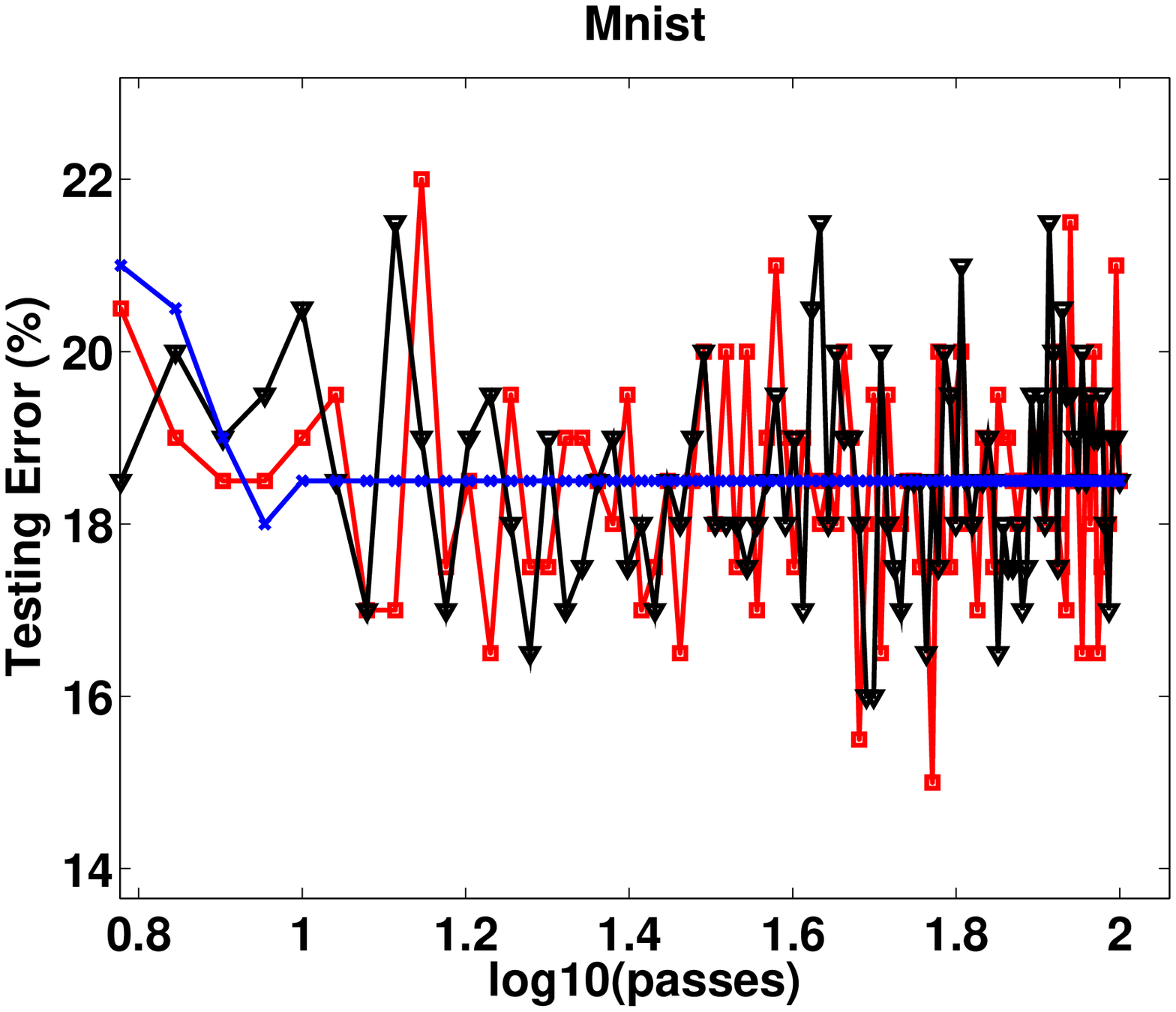}
\includegraphics[width = 1.05in, height = 1.05in]{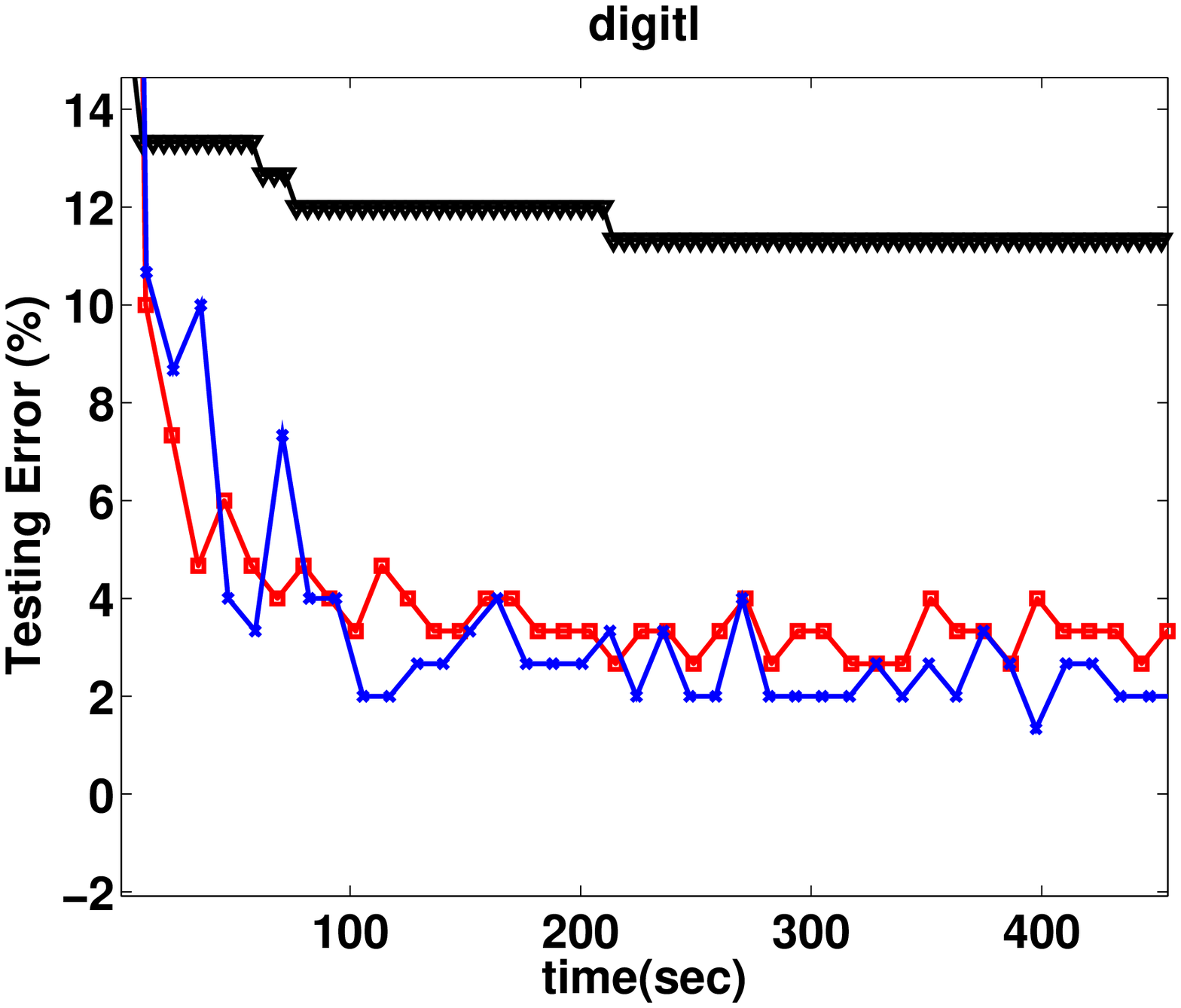}
\includegraphics[width = 1.05in, height = 1.05in]{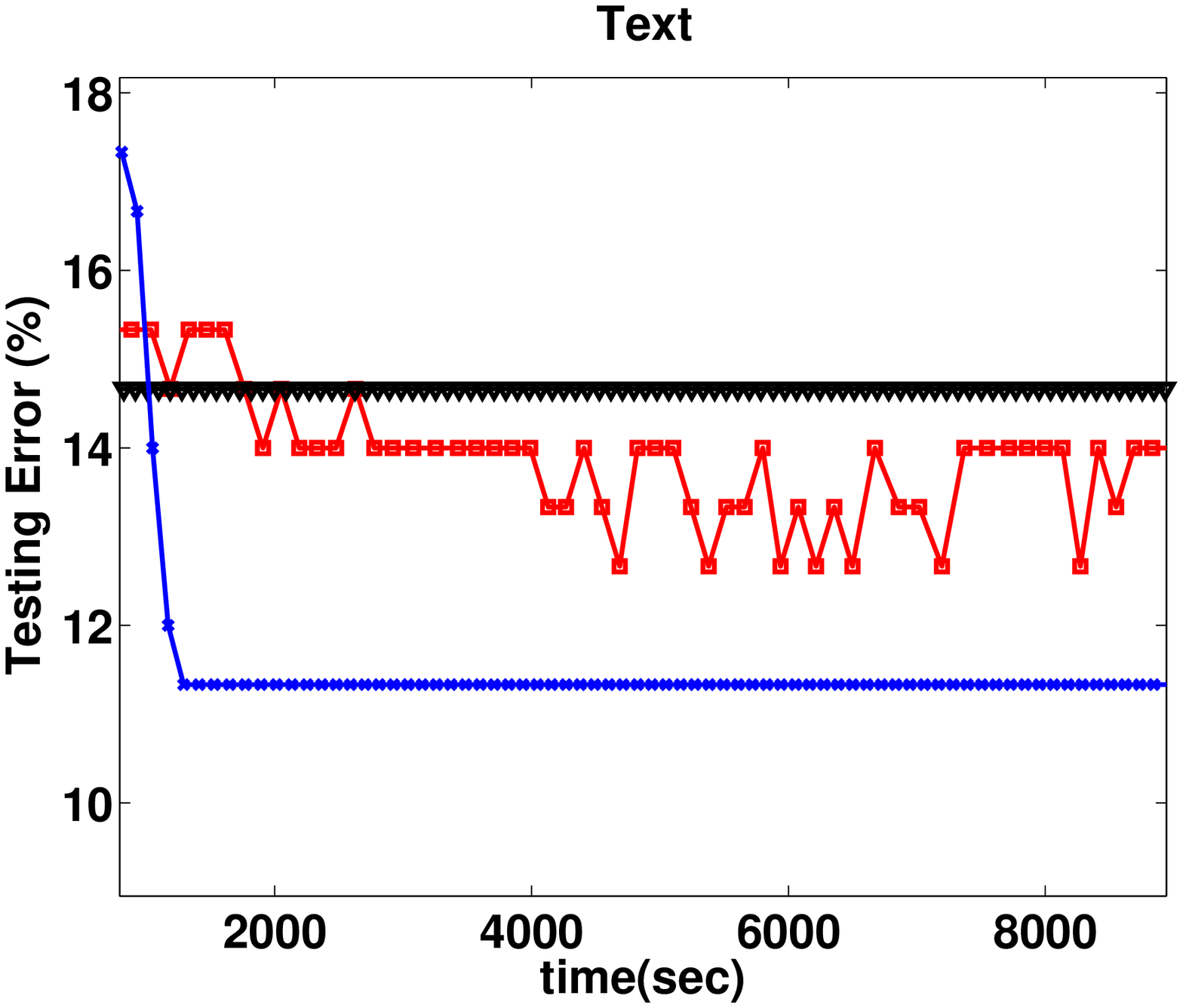} 
\includegraphics[width = 1.05in, height = 1.05in]{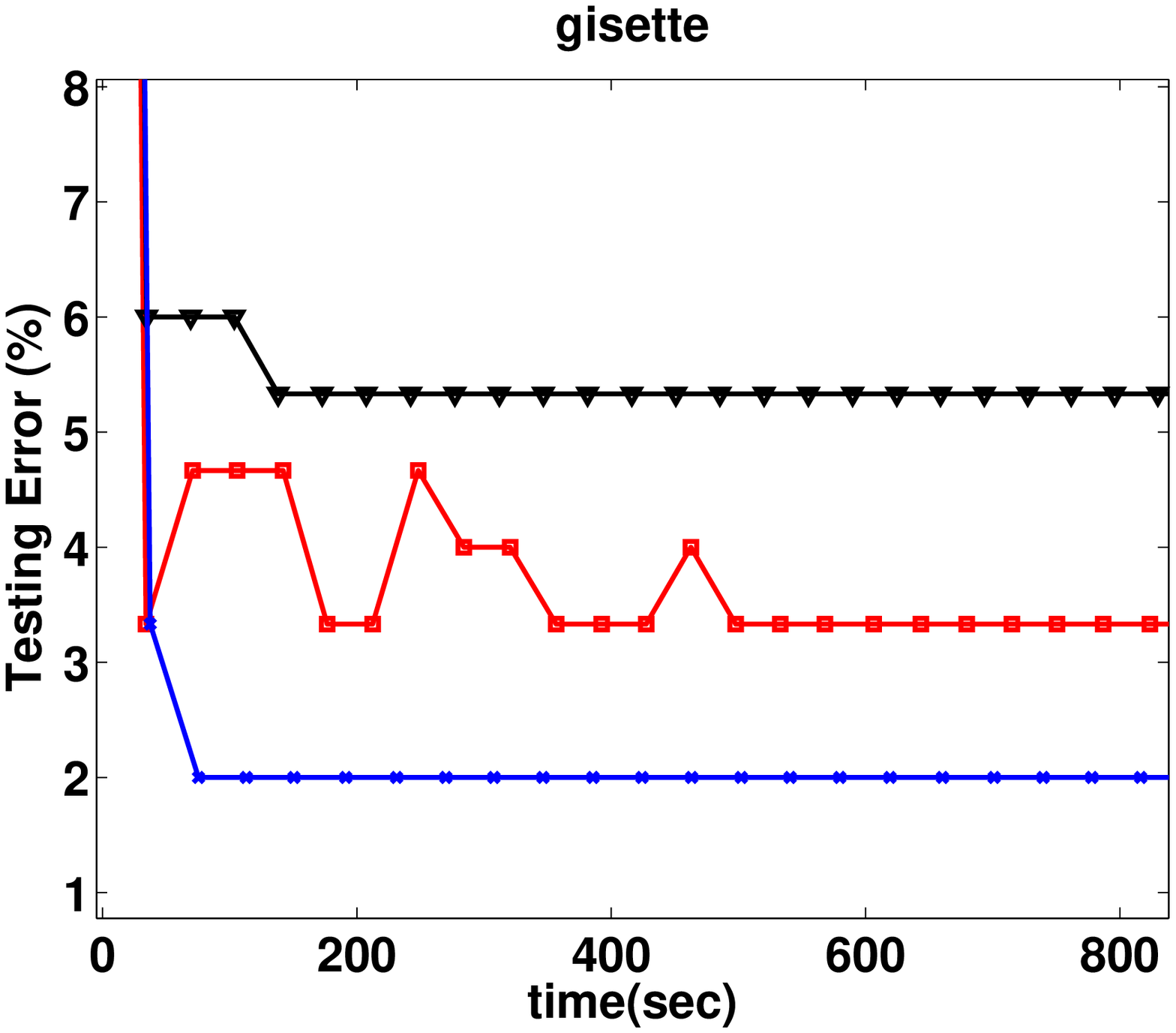} 
\end{tabular}
\caption{A comparison of SGD, ASGD and SBM for $l_2$-regularized logistic regression. \textbf{From top to bottom:} testing log-likelihood and testing error (original and zoomed) vs. passes through the data for data-sets: \textbf{SecStr} ($t = 83679$, $n = 2$, $d = 632$, $\lambda = 10^1$, SGD ($\eta_0 = 10^{-4}$, $m = 10$), ASGD ($\eta_0 = 10^{-1}$, $m = 10$) SBM ($\eta_0 = \frac{1}{t}$, $m = 1$, full-rank)), \textbf{Mnist} ($t = 10000$, $n = 10$, $d = 510$, $\lambda = 10^{-2}$, SGD ($\eta_0 = 10^{-1}$, $m = 100$), ASGD ($\eta_0 = 10^{-1}$, $\tau = 10^{5}$, $m = 100$) SBM ($\eta_0 = \frac{1}{t}$, $m = 1$, full-rank)), \textbf{digitl} ($t = 1500$, $n = 2$, $d = 1448$, $\lambda = 10^1$, SGD ($\eta_0 = 10^{-4}$, $m = 10$), ASGD ($\eta_0 = 10^{-3}$, $\tau = 10^{2}$, $m = 10$) SBM ($\eta_0 = 50\cdot\frac{1}{t}$, $m = 1$, $k = 1$)), \textbf{Text} ($t = 1500$, $n = 2$, $d = 23922$, $\lambda = 10^{1}$, SGD ($\eta_0 = 10^{-4}$, $m = 10$), ASGD ($\eta_0 = 10^{-2}$, $\tau = 10^{3}$, $m = 10$) SBM ($\eta_0 = \frac{1}{t}$, $m = 1$, $k = 1$)) and \textbf{gisette} ($t = 1500$, $n = 2$, $d = 10002$, $\lambda = 10^{0}$, SGD ($\eta_0 = 10^{-3}$, $m = 10$), ASGD ($\eta_0 = 10^{-1}$, $\tau = 10^{1}$, $m = 10$) SBM ($\eta_0 = 100\cdot\frac{1}{t}$, $m = 1$, $k = 1$)).} 
\label{fig:largeresults}
\vspace{-0.2in}
\end{figure}

We next evaluate the performance of the new algorithm empirically. We compare SGD, ASGD and SBM for $l_2$-regularized logistic regression on the Mnist$^{\star}$, gisette$^{\ast}$, SecStr$^{\dagger}$, digitl$^{\dagger}$ and Text$^{\dagger}$ data-sets\footnote{Downloaded 
  from $^{\star}$\textsf{http://yann.lecun.com/exdb/mnist/}, $^{\ast}$\textsf{http://archive.ics.uci.edu/ml/} and $^{\dagger}$\textsf{http://olivier.chapelle.cc/ssl-book/benchmarks.html}}. We show two variants of the SBM algorithm: full-rank (on SecSTR and Mnist) and low-rank (on the remaining data-sets). For the experiments with full-rank SBM, we plot the testing log-likelihood and error versus passes through the data (epoch iterations). For the experiments with low-rank SBM, we plot the likelihood versus cpu time. For SGD and ASGD we tested mini-batches of size from $m = 1$ to $ m = 10$ and chose the best setting. For the Mnist data-set we explored mini-batches of up to $m = 100$ to achieve optimal SGD behavior. For SBM we always simply used $m = 1$. For each data-set, Figure~\ref{fig:largeresults} reports the optimal step size $\eta_0$ for SGD and ASGD and the optimal parameter $\tau$ for ASGD (if it was necessary). For the full-rank version of SBM we always use $\eta_0 = \frac{1}{t}$, however for its low-rank version (where we simply assumed $k = 1$) we tuned $\eta_0$. The chosen value of $\eta_0$ is also reported in Figure~\ref{fig:largeresults}. Clearly, SBM is less prone to over-fitting and achieves higher testing likelihood than SGD and ASGD as well as lower testing error. Simultaneously, SBM exhibits the fastest convergence in terms of the number of passes through the data till convergence. Furthermore, low-rank SBM had the fastest convergence in terms of cpu time, outperforming the leading stochastic first-order methods (SGD and ASGD) as shown in the plots of likelihood over time.

\section{Conclusion}
We have proposed a new stochastic bound majorization method for optimizing the partition function of log-linear models that uses second-order curvature through a global bound (rather than a local Hessian). The method is obtained by applying Sherman-Morrison to the batch update rule to convert it into an iterative summation over the data which can easily be made stochastic by interleaving parameter updates. This (full-rank) stochastic method requires no parameter tuning. A low-rank version of this stochastic update rule makes this effectively second-order method remain linear in the dimensionality of the data. We showed experimentally that the method has significant advantage over the state-of-the-art first-order stochastic methods like SGD and ASGD making majorization competitive in both stochastic and batch settings \cite{JebCho12}. Stochastic bound majorization achieves convergence in fewer iterations, in less computation time (when using the low-rank version), and with better final solutions. Future work will involve providing theoretical guarantees for the method as well as application to deep architectures with cascaded linear combinations of soft-max functions.

\begin{spacing}{0.9}
\bibliographystyle{unsrt}
\small{
\bibliography{nips2013}

\begin{thebibliography}{10}

\bibitem{bottou-98x}
L.~Bottou.
\newblock Online algorithms and stochastic approximations.
\newblock In David Saad, editor, {\em Online Learning and Neural Networks}.
  Cambridge University Press, Cambridge, UK, 1998.

\bibitem{Littlestone:1988:LQI:639961.639994}
N.~Littlestone.
\newblock Learning quickly when irrelevant attributes abound: A new
  linear-threshold algorithm.
\newblock {\em Mach. Learn.}, 2(4):285--318, April 1988.

\bibitem{rosenblatt58a}
F.~Rosenblatt.
\newblock The perceptron: A probabilistic model for information storage and
  organization in the brain.
\newblock {\em Psychological Review}, 65(6):386--408, 1958.

\bibitem{Bottou_stochasticgradient}
L.~Bottou.
\newblock Stochastic gradient learning in neural networks.
\newblock In {\em Proceedings of Neuro-Nˆımes}. 1991.

\bibitem{lecun-98x}
Y.~{Le Cun}, L.~Bottou, G.~B. Orr, and K.-R. M{\"{u}}ller.
\newblock Efficient backprop.
\newblock In {\em Neural Networks, Tricks of the Trade}, Lecture Notes in
  Computer Science LNCS~1524. Springer Verlag, 1998.

\bibitem{DBLP:conf/nips/KrizhevskySH12}
A.~Krizhevsky, I.~Sutskever, and G.~E. Hinton.
\newblock Imagenet classification with deep convolutional neural networks.
\newblock In {\em NIPS}, 2012.

\bibitem{DBLP:conf/interspeech/KingsburySS12}
B.~Kingsbury, T.~N. Sainath, and H.~Soltau.
\newblock Scalable minimum {Bayes} risk training of deep neural network
  acoustic models using distributed {Hessian}-free optimization.
\newblock In {\em INTERSPEECH}, 2012.

\bibitem{Vishwanathan:2006:ATC:1143844.1143966}
S.~V.~N. Vishwanathan, N.~N. Schraudolph, M.~W. Schmidt, and K.~P. Murphy.
\newblock Accelerated training of conditional random fields with stochastic
  gradient methods.
\newblock In {\em ICML}, 2006.

\bibitem{RouxF10}
N.~Le Roux and A.~W. Fitzgibbon.
\newblock A fast natural {Newton} method.
\newblock In {\em ICML}, 2010.

\bibitem{DBLP:conf/nips/WangB12}
C.~Wang and D.~M. Blei.
\newblock Truncation-free online variational inference for {Bayesian}
  nonparametric models.
\newblock In {\em NIPS}, 2012.

\bibitem{ROBINS-MONRO51}
H.~Robbins and S.~Monro.
\newblock A stochastic approximation method.
\newblock {\em Annals of Mathematical Statistics}, 22:400--407, 1951.

\bibitem{DBLP:conf/nips/RouxSB12}
N.~Le Roux, M.~W. Schmidt, and F.~Bach.
\newblock A stochastic gradient method with an exponential convergence rate for
  finite training sets.
\newblock In {\em NIPS}, 2012.

\bibitem{Polyak:1992:ASA:131092.131098}
B.~T. Polyak and A.~B. Juditsky.
\newblock Acceleration of stochastic approximation by averaging.
\newblock {\em SIAM J. Control Optim.}, 30(4):838--855, July 1992.

\bibitem{Tseng:1998:IGM:588881.588930}
P.~Tseng.
\newblock An incremental gradient(-projection) method with momentum term and
  adaptive stepsize rule.
\newblock {\em SIAM J. on Optimization}, 8(2):506--531, February 1998.

\bibitem{DBLP:journals/mp/Nesterov09}
Y.~Nesterov.
\newblock Primal-dual subgradient methods for convex problems.
\newblock {\em Math. Program.}, 120(1):221--259, 2009.

\bibitem{Kesten}
H.~Kesten.
\newblock Accelerated stochastic approximation.
\newblock {\em Annals of Mathematical Statistics}, 29(1):41--59, 1958.

\bibitem{DBLP:journals/siamjo/BlattHG07}
D.~Blatt, A.~O. Hero, and H.~Gauchman.
\newblock A convergent incremental gradient method with a constant step size.
\newblock {\em SIAM Journal on Optimization}, 18(1):29--51, 2007.

\bibitem{Xiao09dualaveraging}
L.~Xiao.
\newblock Dual averaging methods for regularized stochastic learning and online
  optimization.
\newblock In {\em NIPS}, 2009.

\bibitem{DBLP:conf/icml/LeNCLPN11}
Q.~V. Le, J.~Ngiam, A.~Coates, A.~Lahiri, B.~Prochnow, and A.~Y. Ng.
\newblock On optimization methods for deep learning.
\newblock In {\em ICML}, 2011.

\bibitem{DBLP:conf/nips/BottouB07}
L.~Bottou and O.~Bousquet.
\newblock The tradeoffs of large scale learning.
\newblock In {\em NIPS}, 2007.

\bibitem{Schraudolph99localgain}
N.~N. Schraudolph.
\newblock Local gain adaptation in stochastic gradient descent.
\newblock In {\em ICANN}, 1999.

\bibitem{Schraudolph07astochastic}
N.~N. Schraudolph, J.~Yu, and S.~G\"unter.
\newblock A stochastic quasi-{Newton} method for online convex optimization.
\newblock In {\em AISTATS}, 2007.

\bibitem{Amari:2000:AMR:1121517.1121530}
S.-I. Amari, H.~Park, and K.~Fukumizu.
\newblock Adaptive method of realizing natural gradient learning for multilayer
  perceptrons.
\newblock {\em Neural Comput.}, 12(6):1399--1409, June 2000.

\bibitem{Bordes:2009:SCQ:1577069.1755842}
A.~Bordes, L.~Bottou, and P.~Gallinari.
\newblock Sgd-qn: Careful quasi-{Newton} stochastic gradient descent.
\newblock {\em J. Mach. Learn. Res.}, 10:1737--1754, December 2009.

\bibitem{DBLP:conf/icml/Martens10}
J.~Martens.
\newblock Deep learning via {Hessian}-free optimization.
\newblock In {\em ICML}, 2010.

\bibitem{journals/siamsc/FriedlanderS12}
M.~P. Friedlander and M.~W. Schmidt.
\newblock Hybrid deterministic-stochastic methods for data fitting.
\newblock {\em SIAM J. Scientific Computing}, 34(3), 2012.

\bibitem{JebCho12}
T.~Jebara and A.~Choromanska.
\newblock Majorization for {CRFs} and latent likelihoods.
\newblock In {\em NIPS}, 2012.

\bibitem{DBLP:conf/icml/LaffertyMP01}
J.~D. Lafferty, A.~McCallum, and F.~C.~N. Pereira.
\newblock Conditional random fields: Probabilistic models for segmenting and
  labeling sequence data.
\newblock In {\em ICML}, 2001.

\bibitem{DBLP:journals/neco/HintonOT06}
G.~E. Hinton, S.~Osindero, and Y.~W. Teh.
\newblock A fast learning algorithm for deep belief nets.
\newblock {\em Neural Computation}, 18(7):1527--1554, 2006.

\bibitem{Bahl96discriminativetraining}
L.R. Bahl, M.~Padmanabhan, D.~Nahamoo, and P.~S. Gopalakrishnan.
\newblock Discriminative training of {Gaussian} mixture models for large
  vocabulary speech recognition systems.
\newblock In {\em ICASSP}, 1996.

\bibitem{berger97improved}
A.~Berger.
\newblock The improved iterative scaling algorithm: A gentle introduction.
\newblock {\em Technical report, Carnegie Mellon University}, 1997.

\bibitem{LeeuwHeiser2006}
J.~De Leeuw and W.~J. Heiser.
\newblock Convergence of correction matrix algorithms for multidimensional
  scaling.
\newblock {\em chapter Geometric representations of relational data, Mathesis
  Press}, pages 735--752, 1977.

\bibitem{Dempster77maximumlikelihood}
A.~P. Dempster, N.~M. Laird, and D.~B. Rubin.
\newblock Maximum likelihood from incomplete data via the {EM} algorithm.
\newblock {\em Journal of the Royal Statistical Society, Series B},
  39(1):1--38, 1977.

\bibitem{DBLP:conf/icml/SalakhutdinovR03}
R.~Salakhutdinov and S.~T. Roweis.
\newblock Adaptive overrelaxed bound optimization methods.
\newblock In {\em ICML}, 2003.

\end{thebibliography}
}
\end{spacing}

\end{document}